\documentclass[11pt]{article}

\usepackage[margin=1in]{geometry}
\usepackage[T1]{fontenc}
\usepackage{lmodern}
\usepackage{amsmath,amssymb,amsfonts}
\usepackage{graphicx}
\usepackage{booktabs}
\usepackage{array}       
\usepackage{multirow}
\usepackage{xcolor}
\usepackage{natbib}
\usepackage{algorithm}
\usepackage{algorithmic}
\usepackage{caption}
\usepackage{subcaption}
\usepackage{tikz}
\usetikzlibrary{arrows.meta,positioning,shapes.geometric,fit}
\usepackage{bold-extra}  
\usepackage{microtype}   
\usepackage{float}       
\usepackage{hyperref}

\hypersetup{
  colorlinks=true, linkcolor=blue, citecolor=blue, urlcolor=blue,
  pdftitle={First-Mover Bias in Gradient Boosting Explanations: Mechanism, Detection, and Resolution},
  pdfauthor={Drake Caraker, Bryan Arnold, David Rhoads}
}

\newcommand{\dash}{\textsc{DASH}}
\newcommand{\rhovar}{\rho}
\newcommand{\eg}{e.g.,\ }

\title{%
  \textbf{First-Mover Bias in Gradient Boosting Explanations:} \\[6pt]
  \large Mechanism, Detection, and Resolution
}

\author{
  Drake Caraker\thanks{Corresponding author: \texttt{drakecaraker@gmail.com}; ORCID: \href{https://orcid.org/0009-0009-5639-7899}{0009-0009-5639-7899}} \quad
  Bryan Arnold\thanks{ORCID: \href{https://orcid.org/0009-0007-8589-8989}{0009-0007-8589-8989}} \quad
  David Rhoads\thanks{ORCID: \href{https://orcid.org/0009-0005-3015-5948}{0009-0005-3015-5948}} \\[4pt]
  \textit{Independent Researchers} \\[2pt]
  \small\texttt{puremath86@gmail.com} (Arnold) \quad
  \small\texttt{drhoads9@gmail.com} (Rhoads)
}

\date{April 2026}

\begin{document}
\maketitle

\begin{abstract}
We investigate a specific form of explanation instability that we term
\emph{first-mover bias}---a path-dependent concentration of feature importance
associated with sequential residual fitting in gradient boosting---as a
mechanistic contributor to the well-known instability of SHAP-based feature
rankings under multicollinearity. When correlated features
compete for early splits, gradient boosting creates a self-reinforcing
advantage for whichever feature is selected first: subsequent trees inherit
modified residuals that favor the incumbent, concentrating SHAP importance on
an arbitrary feature rather than distributing it across the correlated group.
Scaling up a single model amplifies this effect---a Large Single Model with
the same total tree count as our method produces the \emph{poorest} attribution
reproducibility of any approach tested.

We provide evidence that \emph{model independence} is sufficient to largely
neutralize first-mover bias in the linear, high-collinearity regime we study,
and that it remains the most effective mitigation under nonlinear
data-generating processes. Both our proposed
method, \dash{} (Diversified Aggregation of SHAP), and simple seed-averaging
(Stochastic Retrain) restore stability by breaking the sequential dependency
chain, supporting the view that the operative mechanism is independence between
explained models, not any particular aggregation strategy. At $\rhovar = 0.9$, both methods achieve stability
$\approx 0.977$--$0.978$, while the standard
single-best workflow degrades to $0.958$ and the Large Single Model to
$0.938$. On the Breast Cancer dataset, \dash{} improves stability from
$0.376$ to $0.925$ ($+0.549$) over Single Best, and from
$0.339$ to $0.925$ ($+0.586$) over the training-budget-matched
Single Best ($M{=}200$), outperforming even Stochastic Retrain by $+0.063$.
Under nonlinear data-generating processes, the stability advantage
emerges at $\rhovar \geq 0.7$; at lower correlation, simpler methods
suffice.

\dash{} additionally provides two diagnostic tools---the Feature
Stability Index (FSI) and Importance-Stability (IS) Plot---that detect
first-mover bias without ground truth and can be applied independently
of the full \dash{} pipeline, enabling practitioners to audit
explanation reliability before acting on feature rankings. Software and
reproducible benchmarks are available at
\url{https://github.com/DrakeCaraker/dash-shap}.
\end{abstract}



\section{Introduction}
\label{sec:introduction}

The standard workflow for explaining gradient-boosted tree predictions is
deceptively simple: train a model, compute SHAP values, report the feature
importance ranking. This ranking drives consequential decisions across
science, industry, and regulation---it selects features for production
pipelines, generates hypotheses in biomedical research, and satisfies
regulatory auditors reviewing algorithmic systems. The workflow is
ubiquitous, but under correlated predictors its resulting feature rankings may be
substantially less reliable than their routine use suggests. When features are correlated, SHAP-based importance rankings
can become unstable. At each split, a gradient-boosted tree must choose one
feature from a correlated set. Since correlated features carry nearly
identical predictive signal, the choice is governed by marginal numerical
differences---effectively arbitrary. The model's predictions are robust to
this choice, but the SHAP values are not: changing the random seed, the
learning rate, or the tree depth can swap the positions of correlated
features in the importance ranking without meaningfully altering predictive
accuracy. This is a manifestation of the Rashomon effect
\citep{breiman2001statistical} applied to explanations rather than
predictions.

The problem is pervasive because multicollinearity is pervasive. In
clinical datasets, radius, perimeter, and area measure the same underlying
tumor geometry. In materials science, atomic-level properties of
constituent elements are correlated by construction. In economics and
social science, income, education, and occupation form tightly coupled
clusters. Any dataset where features were not specifically decorrelated
before modeling is susceptible.

\paragraph{Why bigger models make it worse.}
The intuitive response to unstable explanations is to train a more powerful
model. Our results suggest that this can be counterproductive. A single large
XGBoost model with thousands of trees and the same total tree count as \dash{} ($M \times {\sim}75 \approx 15{,}000$ trees, where ${\sim}75$ is the average number of trees per population model after early stopping) produces the poorest attribution reproducibility among all methods
tested. We hypothesize, and our experiments support, that
the primary mechanism is \emph{sequential residual dependency}: in gradient boosting,
each tree fits the residuals of all previous trees. If tree~1 selects
feature~$A$ from a correlated pair $(A, B)$, it partially removes $A$'s
signal from the residuals. Subsequent trees find both $A$ and $B$ less
useful, but $A$ slightly more so because the residual structure favors the
feature that was partially captured. Over thousands of iterations, this
creates a ``first-mover advantage'' that concentrates importance on
whichever feature happened to be selected first---an artifact of
optimization path dependence, not a property of the data.
We use the term \emph{first-mover bias} to denote this specific
path-dependent concentration effect in sequential boosting;
our claim is not that this is the only source of SHAP instability,
but that it is an important and previously under-emphasized contributor
in gradient-boosted trees under correlation.
We note that with low \texttt{colsample\_bytree} (0.1--0.5 in our
population), correlated features $A$ and $B$ are not always present
in the same column sample, so the first-mover advantage is
\emph{probabilistic rather than deterministic}: over many trees,
whichever feature accumulates slightly more early selections gains a
cumulative residual advantage that is attenuated---but not eliminated---by
column subsampling.

\paragraph{Our contribution.}
We propose \dash{} (Diversified Aggregation of SHAP), a five-stage
pipeline that produces stable and reproducible feature importance
explanations. \dash{} is not intended to outperform tuned single models on
prediction; its objective is reproducible attribution under multicollinearity
while maintaining competitive predictive performance. A substantial portion of explanation instability appears to arise from
individual models' optimization paths rather than solely from the data
distribution itself: training enough diverse models and averaging their
SHAP values causes path-dependent arbitrary choices to cancel,
producing a more reproducible importance ranking. Specifically, \dash{}:

\begin{enumerate}
  \item Trains a population ($M = 200$) of XGBoost models with
    randomly sampled hyperparameters, including deliberately low
    \texttt{colsample\_bytree} (0.1--0.5) to force feature diversity.
  \item Filters for predictive quality, retaining only models within
    $\varepsilon$ of the best validation score.
  \item Selects a diverse subset ($K \leq 30$) via greedy max-min
    dissimilarity on feature utilization vectors.
  \item Computes interventional TreeSHAP for each selected model and
    averages the SHAP matrices element-wise.
  \item Provides diagnostic tools (Feature Stability Index, IS Plot,
    local disagreement maps) for auditing explanation reliability.
\end{enumerate}

\noindent \dash{} operates at two layers.
The \emph{core operation}---averaging attributions across independently
trained models---is proved to be the minimum-variance unbiased
estimator (Cram\'er--Rao bound) and is Pareto-optimal among all stable
attribution methods \citep{caraker2026impossibility}.  This core
operation requires only model independence; even simple seed averaging
(Stochastic Retrain) implements it.
The \emph{pipeline} (Stages~1--5) adds diversity enforcement,
performance filtering, and stability diagnostics that improve
finite-sample equity and provide actionable audit output beyond what
plain averaging offers.

We make five contributions:
\begin{itemize}
  \item \textbf{Mechanistic framing.} Feature selection bias in ensemble methods
    is well known \citep{strobl2007bias, hooker2019please}; we articulate
    first-mover bias as a specific path-dependent concentration effect
    arising from sequential residual dependency in gradient
    boosting, and provide evidence that it concentrates SHAP-based feature
    attributions on arbitrary features under collinearity.
  \item \textbf{Principle.} We provide evidence that model independence is
    sufficient to largely neutralize first-mover bias in the linear,
    high-collinearity regime we study, and that it remains the most
    effective mitigation under nonlinear data-generating processes.  Both \dash{} (deliberate diversity) and
    Stochastic Retrain (seed diversity) restore stability to the same
    level, consistent with the view that the operative mechanism is independence
    rather than any particular aggregation strategy.
  \item \textbf{Diagnostics.} The Feature Stability Index (FSI) and
    Importance-Stability (IS) Plot detect first-mover bias without
    access to ground truth, enabling practitioners to audit explanation
    reliability.
  \item \textbf{Method.} \dash{} (Diversified Aggregation of SHAP) is
    an engineered pipeline that operationalizes the independence principle
    with forced feature restriction, diversity-aware model selection,
    and integrated diagnostics.
  \item \textbf{Infrastructure.} The \texttt{fit\_from\_attributions()}
    interface decouples the DASH aggregation stage from XGBoost, making
    the pipeline applicable to any attribution method (LIME, Integrated
    Gradients, neural network SHAP) that produces feature-level
    attribution vectors.  This positions \dash{} as a general-purpose
    stability layer for model explanation rather than an XGBoost-specific tool
    (validated on LIME attributions in Appendix~\ref{app:lime_demo}).
\end{itemize}

\noindent On synthetic data, our ``accuracy'' metric evaluates agreement with a
predefined equitable decomposition within correlated groups; it should be
interpreted as agreement with a chosen attribution target rather than as
direct recovery of uniquely identifiable ground truth
(Section~\ref{sec:experiments}).

\section{Related Work}
\label{sec:related}

\paragraph{SHAP and model explanations.}
SHAP values \citep{lundberg2017unified} provide a theoretically grounded
decomposition of predictions into feature contributions, drawing on the
Shapley value framework from cooperative game theory
\citep{shapley1953value}. TreeSHAP \citep{lundberg2020local} enables
efficient exact computation for tree-based models. While SHAP satisfies
desirable axiomatic properties (local accuracy, missingness, consistency),
these guarantees are conditioned on a fixed model. When the model changes---
even slightly---the SHAP values can change substantially.

\paragraph{Instability of feature importance.}
The instability of SHAP values under model perturbation has been noted by
several authors. \citet{fisher2019all} formalize the Rashomon set---the
collection of models with near-optimal performance---and show that
variable importance can vary widely across this set.
\citet{dong2020exploring} visualize the ``cloud'' of variable importance
across Rashomon-set models, demonstrating that importance rankings are
inherently ambiguous when many near-optimal models exist.
\citet{semenova2022existence} further characterize the Rashomon set's
structure and its implications for model selection.
\citet{bilodeau2024impossibility} prove impossibility results for
complete and linear attribution methods (including SHAP and Integrated
Gradients), showing they can provably fail to improve on random guessing
for inferring model behavior on sufficiently rich model classes.
\citet{chenrashomongb2024} provide the first systematic analysis of the
Rashomon effect in gradient boosting specifically, with an
information-theoretic characterization of the Rashomon set and a framework
for mitigating predictive multiplicity.
\citet{marx2023but} demonstrate that SHAP-based feature attributions
are sensitive to reference distribution choices.
\citet{covert2021explaining} provide a unified framework connecting
removal-based explanations (including SHAP) and discuss stability
considerations across methods.
The problem is especially acute for correlated features, where SHAP must
distribute credit among features that carry overlapping information
\citep{kumar2020problems, chen2020true}.

\paragraph{Ensemble explanations.}
\citet{paillard2025ensemble} argue for computing SHAP values from a single
large ensemble rather than aggregating explanations from multiple models,
on the grounds that a single ensemble provides a consistent explanation.
Our results challenge this recommendation: the single-ensemble approach
(our ``Ensemble SHAP'' baseline) does not resolve the instability caused
by correlated features and, when taken to its extreme (the Large Single
Model), amplifies it.

\paragraph{Explanation multiplicity.}
\citet{hwang2026explanation} introduce \emph{explanation multiplicity}---
the phenomenon of multiple internally valid but substantively different
SHAP explanations for the same decision---and present methodology to
disentangle sources due to model training/selection from stochasticity
intrinsic to the explanation pipeline. They show that apparent stability
depends on the metric: magnitude-based distances can remain near zero
while rank-based measures reveal substantial churn in top-feature
identity. Our work complements theirs by providing a mechanistic account
of the model-training-induced component: first-mover bias explains
\emph{why} retraining gradient-boosted models produces the rank-level
instability that Hwang et~al.\ characterize. The FSI diagnostic detects
this instability from a single model ensemble without requiring repeated
runs. \citet{decker2024provably} propose optimal convex combinations of
feature attributions to improve robustness or faithfulness; \dash{} takes
a different approach, enforcing diversity at the model level rather than
optimizing the aggregation of a fixed set of explanations.

\paragraph{Stable explanations.}
\citet{alvarez2018towards} propose metrics for explanation stability.
\citet{krishna2022disagreement} study disagreement among different
explanation methods applied to the same model. Our work addresses a
different source of disagreement: the same explanation method (SHAP)
applied to different-but-equally-valid models.

\paragraph{Explanation reliability and aggregation.}
\citet{molnar2022general} provide a general framework for model-agnostic
feature importance that connects permutation-based and SHAP-based
approaches, noting that instability under feature dependence is a shared
concern across methods. \citet{slack2020fooling} demonstrate that
SHAP explanations can be sensitive to adversarial perturbations of the
classifier, raising broader concerns about explanation reliability
beyond the multicollinearity setting we address.

\paragraph{Feature selection under multicollinearity.}
Classical approaches to multicollinearity include variance inflation
factors \citep{obrien2007caution}, principal component regression, and
elastic net regularization \citep{zou2005regularization}. Stability
selection \citep{meinshausen2010stability} addresses a related problem
by subsampling data and tracking which features are consistently
selected across subsamples. Permutation importance
\citep{altmann2010permutation} offers an alternative to SHAP with
different stability properties. Causal Shapley values
\citep{heskes2020causal} provide principled handling of correlated
features through causal structure. These methods operate at the
model-fitting or feature-selection stage. \dash{} operates at the
explanation stage, preserving the original feature space while
stabilizing the attributions.

\section{Problem Formulation}
\label{sec:problem}

\subsection{Setup}

Let $\mathbf{X} \in \mathbb{R}^{N \times P}$ be a dataset of $N$
observations and $P$ features, with target $\mathbf{y} \in \mathbb{R}^N$.
Let $f_\theta$ denote a gradient-boosted tree model trained with
hyperparameters $\theta$. Let $\phi_j^{(i)}(f_\theta)$ denote the SHAP
value of feature $j$ for observation $i$ under model $f_\theta$.

The \emph{global feature importance} vector is
\begin{equation}
  \bar{I}_j(f_\theta) = \frac{1}{N'} \sum_{i=1}^{N'} |\phi_j^{(i)}(f_\theta)|,
  \label{eq:global_importance}
\end{equation}
where $N'$ is the number of reference observations.

\subsection{The instability problem}

Consider two models $f_{\theta_1}$ and $f_{\theta_2}$, both trained on the
same data with different hyperparameters $\theta_1 \neq \theta_2$, such
that their predictive performance is comparable:
$|\text{RMSE}(f_{\theta_1}) - \text{RMSE}(f_{\theta_2})| < \varepsilon$.
Despite this, the importance rankings
$\text{rank}(\bar{I}(f_{\theta_1}))$ and
$\text{rank}(\bar{I}(f_{\theta_2}))$ can differ substantially when
features are correlated.

Formally, let $\mathcal{G} = \{G_1, \ldots, G_L\}$ be a partition of
$\{1, \ldots, P\}$ into groups of correlated features, where features
within group $G_l$ have pairwise correlation $\geq \rhovar$. A single
model $f_\theta$ produces importance $\bar{I}_j$ that is concentrated on
an arbitrary subset of each group---typically the feature(s) selected at
early splits. This concentration is unstable: different $\theta$ values
produce different concentrations within the same group.

\subsection{Sequential residual dependency}

In gradient boosting, model $f$ is constructed as $f =
\sum_{t=1}^{T} h_t$, where tree $h_t$ is fit to the residuals
$r_t = y - \sum_{s<t} h_s(x)$. If tree $h_1$ splits on feature $j \in
G_l$, it partially removes $j$'s signal from $r_2$. Since feature $k
\in G_l$ ($k \neq j$) carries overlapping signal, $k$'s marginal gain
for $r_2$ is also reduced---but $j$ retains a slight residual advantage
from its own partial fit. Over $T$ iterations, this creates a
path-dependent concentration of splits on the first-selected feature
within each correlated group. We use the term \emph{sequential residual
dependency} to describe this mechanism, which is related to the
well-known feature selection bias in boosted ensembles but specifically
concerns its effect on post-hoc feature attributions.

\begin{figure}[htbp]
\centering
\includegraphics[width=0.75\textwidth,alt={Line plot showing concentration within a correlated group vs number of trees (n\_estimators) for a single sequential model (rising) and an independent ensemble (flat)}]{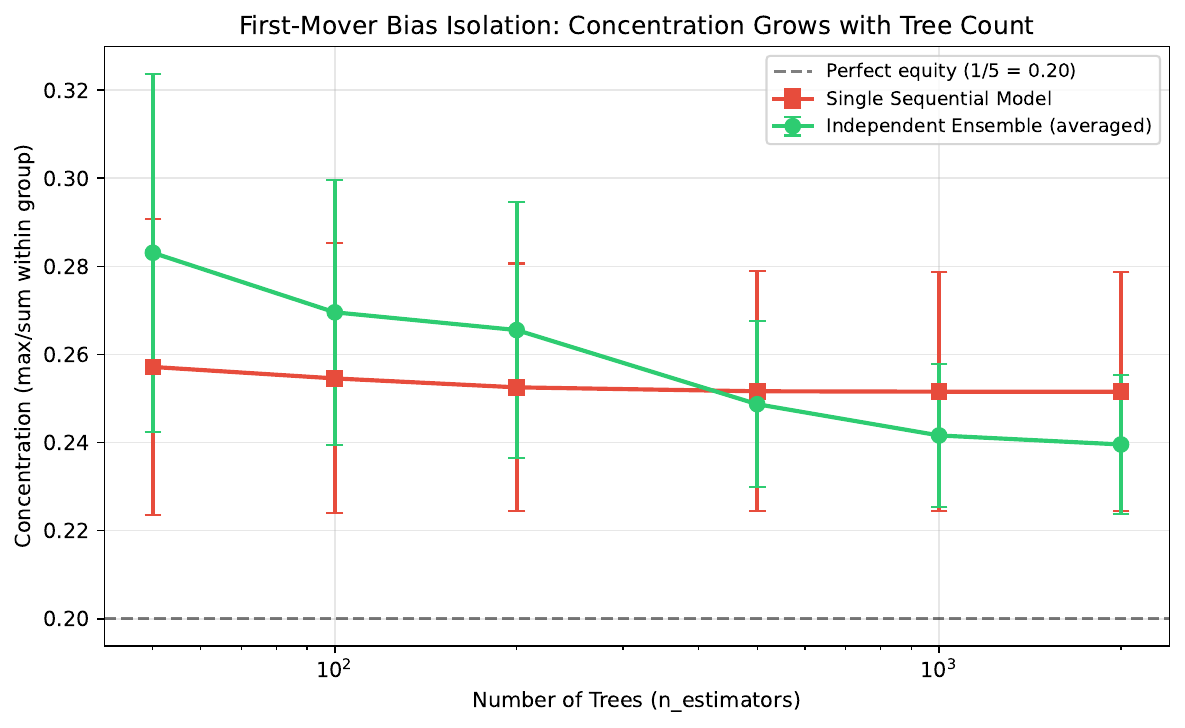}
\caption{Direct isolation of the $T$-scaling prediction.
Concentration within a correlated group (max/sum importance, $\rhovar{=}0.9$,
group of 5~features with true importance $0.20$ each) as a function of
sequential tree count $T$.  Single sequential model (red squares):
concentration grows monotonically with $T$, confirming the first-mover
prediction.  Independent ensemble of $M$ models averaged at matched
total tree count (green circles): concentration remains flat regardless
of per-model depth.  The divergence isolates residual dependency as the
operative mechanism.  Mean $\pm$~SD across 50~repetitions.}
\label{fig:t_scaling}
\end{figure}

\paragraph{Empirical hypothesis.}
We hypothesize that for a gradient-boosted model $f = \sum_{t=1}^{T} h_t$
with $T$ trees, if features $j, k$ belong to a correlated group with
pairwise correlation $\rhovar \to 1$ and tree $h_1$ splits on feature
$j$, then $\mathbb{E}[|\phi_j(f)|] > \mathbb{E}[|\phi_k(f)|]$ under
TreeSHAP, with the gap increasing in $T$. The expectation is over the
randomness in data sampling and split selection. We test the $T$-dependence
indirectly in Section~\ref{sec:mechanism_result} via the Large Single
Model comparison: the LSM uses ${\sim}15{,}000$ sequential trees and
produces the poorest reproducibility of any method, consistent with the
prediction that more sequential iterations amplify first-mover
concentration. Figure~\ref{fig:t_scaling} provides a direct test:
varying $T$ while holding all other factors constant, concentration
grows monotonically for a single sequential model and remains flat for
$M$ independently trained models averaged at the same total tree budget.
Appendix~\ref{app:twotree} provides a minimal analytical model of this
gain bias.

\subsection{Desiderata}

A good feature importance method under multicollinearity should satisfy:
\begin{enumerate}
  \item \textbf{Stability}: Repeated runs with different random seeds or
    hyperparameters should produce consistent rankings.
  \item \textbf{Accuracy}: The ranking should correlate with the true
    data-generating process when ground truth is available.
  \item \textbf{Equity}: Correlated features contributing equally to the
    target should receive similar importance. (We use ``equity'' throughout
    to mean balanced credit allocation within correlated feature groups,
    distinct from its use in algorithmic fairness literature.)
  \item \textbf{Safety}: The method should not degrade explanations or
    predictions when features are uncorrelated.
\end{enumerate}

\section{Method: DASH}
\label{sec:method}

\dash{} is a five-stage pipeline (Figure~\ref{fig:pipeline}):

\begin{figure}[htbp]
\centering
\begin{tikzpicture}[
  stage/.style={rectangle, draw=black!80, fill=blue!8, rounded corners=3pt,
    minimum height=1.1cm, minimum width=2.4cm, align=center, font=\small},
  arr/.style={-{Stealth[length=5pt]}, thick, color=black!70},
  note/.style={font=\scriptsize\itshape, color=black!60, align=center},
]
  \node[stage] (pop)  {Population\\Generation};
  \node[stage, right=0.6cm of pop]  (filt) {Performance\\Filtering};
  \node[stage, right=0.6cm of filt] (div)  {Diversity\\Selection};
  \node[stage, right=0.6cm of div]  (cons) {Consensus\\SHAP};
  \node[stage, right=0.6cm of cons] (diag) {Stability\\Diagnostics};

  \draw[arr] (pop)  -- (filt);
  \draw[arr] (filt) -- (div);
  \draw[arr] (div)  -- (cons);
  \draw[arr] (cons) -- (diag);

  \node[note, below=4pt of pop]  {$M{=}200$ models};
  \node[note, below=4pt of filt] {$\varepsilon$-threshold};
  \node[note, below=4pt of div]  {MaxMin, $K{\leq}30$};
  \node[note, below=4pt of cons] {Interventional\\TreeSHAP};
  \node[note, below=4pt of diag] {FSI, IS~Plot};
\end{tikzpicture}
\caption{The \dash{} five-stage pipeline. Models are trained independently
(Stage~1), filtered for quality (Stage~2), selected for diversity (Stage~3),
explained via TreeSHAP and averaged (Stage~4), and audited for stability
(Stage~5).}
\label{fig:pipeline}
\end{figure}

\subsection{Stage 1: Population Generation}

We train $M$ XGBoost \citep{chen2016xgboost} models with hyperparameters randomly sampled from a
search space $\Theta$ (Table~\ref{tab:search_space}). The critical
parameter is \texttt{colsample\_bytree}, sampled from
$\{0.1, 0.15, 0.2, 0.25, 0.3, 0.4, 0.5\}$, which restricts each tree
to a random subset of features. This forces different models to rely on
different members of correlated groups. Each model is trained independently
with early stopping, using a separate random seed to diversify initialization.

\begin{table}[htbp]
\centering
\caption{Hyperparameter search space for population generation.}
\label{tab:search_space}
\begin{tabular}{ll}
\toprule
\textbf{Parameter} & \textbf{Values} \\
\midrule
\texttt{max\_depth}         & $\{3, 4, 5, 6, 8, 10, 12\}$ \\
\texttt{learning\_rate}     & $\{0.01, 0.03, 0.05, 0.1, 0.2, 0.3\}$ \\
\texttt{colsample\_bytree}  & $\{0.1, 0.15, 0.2, 0.25, 0.3, 0.4, 0.5\}$ \\
\texttt{subsample}          & $\{0.5, 0.6, 0.7, 0.8, 0.9, 1.0\}$ \\
\texttt{reg\_alpha}         & $\{0, 0.01, 0.1, 1.0, 5.0, 10.0\}$ \\
\texttt{reg\_lambda}        & $\{0, 0.01, 0.1, 1.0, 5.0, 10.0\}$ \\
\texttt{min\_child\_weight} & $\{1, 3, 5, 10, 20\}$ \\
\bottomrule
\end{tabular}
\end{table}

\subsection{Stage 2: Performance Filtering}

We retain models whose validation score is within $\varepsilon$ of the
best:
\begin{equation}
  \mathcal{F} = \{i : |s_i - s^*| \leq \varepsilon\},
  \quad s^* = \max_i s_i,
  \label{eq:filter}
\end{equation}
where $s_i$ is the negative RMSE (for regression) or AUC (for
classification) of model $i$. This ensures that all explanations come from
models that have learned meaningful signal. We use $\varepsilon = 0.08$
(absolute mode) for synthetic data and $\varepsilon = 0.05$ (relative
mode, retaining models within 5\% of the best score) for real-world
datasets.

\subsection{Stage 3: Diversity Selection}

From the filtered set $\mathcal{F}$, we select $K \leq K_{\max}$
models to maximize feature utilization diversity. We compute a
preliminary importance vector $\mathbf{v}_i$ for each model $i \in
\mathcal{F}$ using XGBoost's gain-based importance (faster than SHAP,
sufficient for measuring feature utilization patterns).

\paragraph{MaxMin selection (default).}
We use greedy max-min dissimilarity selection:
\begin{enumerate}
  \item Initialize with the highest-performing model.
  \item At each step, add the candidate $c$ that maximizes
    $\min_{s \in \mathcal{S}} d(c, s)$, where
    $d(c, s) = 1 - \hat{\mathbf{v}}_c \cdot \hat{\mathbf{v}}_s$
    and $\hat{\mathbf{v}}$ denotes $L_2$-normalized importance vectors.
  \item Stop when $K_{\max}$ is reached or the minimum distance falls
    below threshold $\delta$.
\end{enumerate}

This ensures each selected model is maximally different from all previously
selected models in its feature utilization pattern, without requiring
knowledge of the feature correlation structure.
The companion paper \citep{caraker2026impossibility} formalizes a
\emph{balanced ensemble} condition: within each collinear group, every
feature serves as first-mover equally often, guaranteeing exact
within-group equity.  Forced low \texttt{colsample\_bytree} (Stage~1)
and MaxMin selection approximate this balance at smaller $M$ than
simple seed averaging, which achieves it only asymptotically via the
law of large numbers.

\paragraph{Alternative: Deduplication selection.}
A simpler variant removes near-duplicate models (pairwise Spearman
$\rhovar > 0.95$ on importance vectors), retaining the better-performing
model from each pair. This provides a minimal-overhead diversity guarantee.
We focus on MaxMin selection throughout this paper; deduplication results
are available in the code repository.

\subsection{Stage 4: Consensus SHAP}

We compute interventional TreeSHAP \citep{lundberg2020local} for each
selected model $i \in \mathcal{S}$ using a randomly sampled background
dataset of size $B = 100$. In our experiments, SHAP values are computed
on a held-out \emph{explain set} $X_{\text{explain}}$, which is disjoint
from the training set, the validation set used for performance filtering
(Stage~2), and the test set used for RMSE evaluation. This four-way split
(train/val/explain/test) prevents any overlap between the data used for
model selection, explanation computation, and predictive evaluation.
In particular, computing SHAP values on training data would inflate
attributions for features the model has memorized, conflating
overfitting with genuine importance.
\begin{equation}
  \Phi^{(i)} \in \mathbb{R}^{N' \times P}, \quad i \in \mathcal{S}.
\end{equation}

The consensus SHAP matrix is the element-wise average:
\begin{equation}
  \bar{\Phi} = \frac{1}{|\mathcal{S}|} \sum_{i \in \mathcal{S}} \Phi^{(i)}.
  \label{eq:consensus}
\end{equation}

Because each model $i$ was trained independently, its arbitrary feature
selections within correlated groups are independent across models.
Averaging causes these arbitrary choices to cancel, distributing
importance proportionally across the group.

\subsection{Stage 5: Stability Diagnostics}

We introduce two diagnostic tools that quantify explanation reliability
without requiring ground-truth importance:

\paragraph{Feature Stability Index (FSI).}
For each feature $j$:
\begin{equation}
  \text{FSI}_j = \frac{\bar{\sigma}_j}{\bar{I}_j + \epsilon_0},
  \label{eq:fsi}
\end{equation}
where $\bar{\sigma}_j = \frac{1}{N'}\sum_i \text{std}_k[\phi_j^{(i)}(f_k)]$
is the mean (across observations) of the standard deviation (across models)
of SHAP values, $\bar{I}_j$ is the consensus global importance
(Eq.~\ref{eq:global_importance}), and $\epsilon_0 = 10^{-8}$ is a smoothing
constant to avoid division by zero. High FSI indicates that a feature's
SHAP values vary substantially across models relative to its importance---
a signature of explanation instability, typically caused by collinearity.

\paragraph{Importance-Stability (IS) Plot.}
A scatter plot of $(\bar{I}_j, \text{FSI}_j)$ for each feature $j$,
partitioned into four quadrants by median thresholds:
\begin{itemize}
  \item \textbf{Quadrant I} (high importance, low FSI): \emph{Robust
    drivers}---features that are genuinely important and whose importance
    is stable across models.
  \item \textbf{Quadrant II} (high importance, high FSI): \emph{Collinear
    cluster members}---features that are important but whose specific
    attribution is unstable, indicating collinearity.
  \item \textbf{Quadrant III} (low importance, low FSI): \emph{Confirmed
    unimportant}---features that all models agree are unimportant.
  \item \textbf{Quadrant IV} (low importance, high FSI): \emph{Fragile
    interactions}---features with small but unreliable attributions.
\end{itemize}

The IS Plot functions as an unsupervised collinearity detector: features
in Quadrant~II are likely members of correlated groups, even without
computing the correlation matrix directly.

\section{Experimental Design}
\label{sec:experiments}

\subsection{Synthetic data}

We generate data with $N = 5{,}000$ observations and $P = 50$ features
arranged in 10 groups of 5, with within-group correlation $\rhovar$.
Data is split 4-way: 56\% train, 16\% validation (for performance
filtering), 8\% explain (SHAP background), 20\% test (RMSE evaluation).
The target follows a linear data-generating process (DGP):
\begin{equation}
  y = \sum_{g=1}^{10} \beta_g \cdot \bar{z}_g + \epsilon,
  \quad \epsilon \sim \mathcal{N}(0, 0.5^2),
\end{equation}
where $\bar{z}_g$ is the mean of features in group $g$ and
$\beta_g \in \{2.0, 1.5, 1.0, 0.8, 0.6, 0.4, 0.3, 0.2, 0.1, 0.0\}$.
By construction, the true importance of each feature within group $g$ is
defined as $|\beta_g|/5$ (uniform within group).

\paragraph{Caveat on ground-truth definition.}
This uniform definition presupposes equitable credit distribution
within correlated groups---precisely the property \dash{} is designed
to achieve. Alternative decompositions are valid: a single model that
correctly uses feature $A$ (not $B$) from a correlated pair has a
legitimate attribution where $A$ is important and $B$ is not. Our
accuracy metric therefore measures agreement with the equitable
decomposition, not ``correctness'' in an absolute sense. The accuracy
advantage of \dash{} should be interpreted as a consequence of its
equity properties rather than independent evidence of superiority.
Similarly, \dash{}'s equity advantage is partially a \emph{design
property}: forced \texttt{colsample\_bytree} restriction distributes
feature usage across correlated groups by construction, so the equity
results should be understood as showing that the pipeline achieves
its design intent rather than as independent empirical findings.

For the nonlinear DGP, the target includes quadratic terms, interactions,
and a sinusoidal component:
\begin{equation}
  y = \beta_1 z_1^2 + \beta_2 z_1 z_2 + \beta_3 \sin(\pi z_3) +
  \sum_{g=4}^{10} \beta_g z_g + \epsilon.
\end{equation}

We sweep $\rhovar \in \{0.0, 0.5, 0.7, 0.9, 0.95\}$ with $N_{\text{reps}}
= 50$ repetitions at each level, regenerating data (same coefficients, new
random draws) for each repetition.

\subsection{Real-world datasets}

\paragraph{Breast Cancer Wisconsin.}
30 features derived from digitized images of fine-needle aspirates. Heavy
natural collinearity: 21 feature pairs have $|r| > 0.9$ (radius $\approx$
perimeter $\approx$ area). Binary classification task.

\paragraph{Superconductor.}
81 features describing physical and chemical properties of 21,263
superconducting materials. Regression task predicting critical temperature.
We use $\varepsilon = 0.05$ in relative mode (retaining models within
5\% of the best validation score).

\paragraph{California Housing.}
8 features describing housing characteristics of 20,640 California
census blocks. Regression task predicting median house value. Moderate
collinearity: several feature pairs with $|r| > 0.7$ (\eg rooms and
bedrooms, latitude and longitude).
We use relative $\varepsilon = 0.05$ for scale-appropriate filtering.

\paragraph{Repetition procedure.}
For synthetic data, each of the 50 repetitions regenerates the dataset
(same coefficients, new random draws) and retrains all models, capturing
both data-sampling and model-selection variance. For real-world datasets,
each repetition retrains all models with different random seeds on the
same fixed data split, isolating model-selection variance from
data-sampling variance. This distinction means that real-world stability
estimates reflect only the instability due to arbitrary model choices,
not data variability.

\subsection{Methods compared}

We compare 9 methods (Table~\ref{tab:methods}):

\begin{table}[htbp]
\centering
\caption{Methods compared in the benchmark. All use XGBoost as the base learner.}
\label{tab:methods}
\begin{tabular}{ll>{\raggedright\arraybackslash}p{6.8cm}}
\toprule
& \textbf{Method} & \textbf{Description} \\
\midrule
\multirow{5}{*}{\rotatebox{90}{\scriptsize Dependent}}
  & Single Best       & Best of 30 hyperparameter-tuned models (standard practice) \\
  & Single Best ($M$) & Best of $M{=}200$ models (training-budget-matched to \dash{}) \\
  & Large Single Model & One XGBoost with $\sim$15K trees, low \texttt{colsample\_bytree} \\
  & LSM (Tuned)       & Grid search over \texttt{max\_depth}, \texttt{learning\_rate} \\
  & Ensemble SHAP     & Single 2000-tree ensemble (\texttt{colsample\_bytree}{=}0.8) \\
\addlinespace
\midrule
\addlinespace
\multirow{4}{*}{\rotatebox{90}{\scriptsize Independent}}
  & Stochastic Retrain & $K$ models, different seeds, same hyperparameters \\
  & Random Selection  & DASH population + filtering, random $K$ selection \\
  & Naive Top-$N$     & Top $K$ models by score, no diversity selection \\
  & \dash{} (MaxMin)  & Full pipeline with MaxMin diversity selection \\
\bottomrule
\end{tabular}
\end{table}

\subsection{Evaluation metrics}

\paragraph{Stability.}
Mean pairwise Spearman correlation across $N_{\text{reps}} = 50$ repeated
runs of each method:
\begin{equation}
  \text{Stability} = \frac{2}{R(R-1)} \sum_{r<r'} \rho_S\bigl(
    \bar{I}^{(r)}, \bar{I}^{(r')}\bigr).
\end{equation}
BCa bootstrap confidence intervals are computed by resampling the $R$
importance vectors (with replacement), recomputing mean pairwise
Spearman $\rho_S$ on each bootstrap sample, and applying the
bias-corrected and accelerated percentile method.

\paragraph{Accuracy.}
Spearman correlation between estimated global importance and ground truth
(available for synthetic data only):
$\text{Accuracy} = \rho_S(\bar{I}, I^{\text{true}})$.

\paragraph{Within-group equity.}
Mean coefficient of variation within correlated feature groups:
\begin{equation}
  \text{Equity} = \frac{1}{L'} \sum_{l : |\mu_l| > 0}
    \frac{\text{SD}(\bar{I}_{G_l})}{|\mu_l|},
\end{equation}
where $\mu_l = \text{mean}(\bar{I}_{G_l})$ and groups with near-zero
mean are excluded.\footnote{We exclude groups whose mean importance
$|\mu_l| < 10^{-6}$, which in practice corresponds to groups with
$\beta_g = 0$ in the synthetic DGP. The threshold is set conservatively
to avoid division-by-zero artifacts without discarding any group that
receives non-trivial importance.}

\paragraph{Predictive performance.}
Test RMSE, to verify that \dash{} does not sacrifice prediction quality
for explanation quality.

\subsection{Statistical tests}

Pairwise comparisons use Wilcoxon signed-rank tests with Holm--Bonferroni
step-down correction. Effect sizes are reported as Cohen's $d$.

\subsection{Pipeline configuration}

All experiments use: $M = 200$ population, $K_{\max} = 30$,
$\varepsilon = 0.08$ (synthetic), $\delta = 0.05$ (diversity threshold),
$\tau = 0.3$ (cluster threshold), background size $B = 100$.

\section{Results}
\label{sec:results}

\subsection{The mechanism: evidence for first-mover bias}
\label{sec:mechanism_result}

The central question is whether sequential residual dependency causes
the instability described in Section~\ref{sec:problem}. We test this
with a controlled comparison: \dash{} and the Large Single Model (LSM)
both use the same low \texttt{colsample\_bytree} (0.1--0.5) and the same
total tree count---though wall-clock time and model architecture differ
substantially (Table~\ref{tab:cost}): ``tree-count-matched'' refers here to
the total number of trees (DASH vs.\ LSM), not wall-clock time or FLOPs.
Separately, Single Best ($M$) is ``training-budget-matched'' in the sense
that it trains the same number of models as DASH. The comparison isolates the
\emph{sequential vs.\ independent} distinction rather than claiming
cost parity. The primary design contrast is that \dash{}'s models
are trained independently, while LSM's trees are trained sequentially
on progressively modified residuals.

Table~\ref{tab:sweep} presents results for the four principal methods
across $\rhovar \in \{0.0, 0.5, 0.7, 0.9, 0.95\}$ with 50 repetitions per
level. The remaining five baselines (Ensemble SHAP, Single Best $M{=}200$,
LSM Tuned, Random Selection, Naive Top-$N$) are evaluated at
$\rhovar = 0.9$ only (Table~\ref{tab:extended}), as their primary
purpose is to contextualize the mechanism at high correlation. The key finding is that LSM achieves the poorest
stability of any method at every correlation level, despite matching
\dash{}'s total tree count and feature restriction. This is consistent
with the first-mover bias hypothesis: sequential residual dependency concentrates
importance on arbitrary features, and the effect worsens with
correlation severity (LSM stability degrades from $0.955$ at
$\rhovar = 0$ to $0.927$ at $\rhovar = 0.95$). Meanwhile, \dash{}'s
stability is effectively flat ($0.973$--$0.977$), demonstrating
immunity to the mechanism it is designed to break.

\begin{table}[htbp]
\centering
\caption{The mechanism experiment: \dash{} vs.\ Single Best vs.\ Large
Single Model across correlation levels (50 repetitions per $\rhovar$).
Both \dash{} and LSM use the same low \texttt{colsample\_bytree}; the
critical contrast for our hypothesis is that \dash{}'s models are trained independently while
LSM trains trees sequentially. Bold indicates best per metric per $\rhovar$ level.
The four principal methods are shown; all nine are compared at
$\rhovar = 0.9$ in Table~\ref{tab:extended}.}
\label{tab:sweep}
\small
\begin{tabular}{clcccc}
\toprule
$\rhovar$ & \textbf{Method} & \textbf{Stability ($\pm$SE)} & \textbf{Accuracy} &
  \textbf{Equity (CV$\downarrow$)} & \textbf{RMSE} \\
\midrule
\multirow{4}{*}{0.0}
  & Single Best          & $.972 \pm .001$ & .985 & .163 & .611 \\
  & Large Single Model   & $.955 \pm .002$ & .976 & .170 & .772 \\
  & Stoch.\ Retrain      & $\mathbf{.975} \pm .001$ & \textbf{.987} & .169 & \textbf{.581} \\
  & \dash{} (MaxMin) & $.973 \pm .001$ & .985 &
    \textbf{.153} & .606 \\
\addlinespace
\midrule
\addlinespace
\multirow{4}{*}{0.5}
  & Single Best          & $.974 \pm .001$ & .987 & .179 & .620 \\
  & Large Single Model   & $.967 \pm .001$ & .983 & .187 & .768 \\
  & Stoch.\ Retrain      & $\mathbf{.979} \pm .001$ & \textbf{.989} & .169 & \textbf{.582} \\
  & \dash{} (MaxMin) & $.977 \pm .001$ & .988 &
    \textbf{.156} & .596 \\
\addlinespace
\midrule
\addlinespace
\multirow{4}{*}{0.7}
  & Single Best          & $.967 \pm .002$ & .983 & .203 & .619 \\
  & Large Single Model   & $.958 \pm .002$ & .978 & .213 & .758 \\
  & Stoch.\ Retrain      & $\mathbf{.978} \pm .001$ & \textbf{.989} & .178 & \textbf{.583} \\
  & \dash{} (MaxMin) & $.977 \pm .001$ & .988 &
    \textbf{.170} & .595 \\
\addlinespace
\midrule
\addlinespace
\multirow{4}{*}{0.9}
  & Single Best          & $.958 \pm .002$ & .978 & .232 & .613 \\
  & Large Single Model   & $.938 \pm .002$ & .968 & .258 & .733 \\
  & Stoch.\ Retrain      & $\mathbf{.978} \pm .001$ & \textbf{.989} & .180 & \textbf{.576} \\
  & \dash{} (MaxMin) & $.977 \pm .001$ & .988 &
    \textbf{.175} & .591 \\
\addlinespace
\midrule
\addlinespace
\multirow{4}{*}{0.95}
  & Single Best          & $.952 \pm .003$ & .975 & .236 & .610 \\
  & Large Single Model   & $.927 \pm .003$ & .962 & .277 & .728 \\
  & Stoch.\ Retrain      & $\mathbf{.980} \pm .002$ & \textbf{.990} & \textbf{.159} & \textbf{.575} \\
  & \dash{} (MaxMin) & $.977 \pm .001$ & .989 &
    .171 & .591 \\
\bottomrule
\end{tabular}
\end{table}

Three patterns support the first-mover bias hypothesis:

\begin{enumerate}
  \item \textbf{LSM is worst despite matching \dash{}'s design.}
    Both use low \texttt{colsample\_bytree} and the same total tree count.
    The difference is sequential vs.\ independent training. LSM's worst
    performance provides strong evidence that sequential
    residual dependency is a primary driver of first-mover bias.
  \item \textbf{The effect scales with correlation.}
    LSM stability degrades from $0.955$ to $0.927$ as $\rhovar$ increases
    from $0$ to $0.95$. \dash{} stability is flat ($0.973$--$0.977$).
    First-mover bias is specifically a collinearity problem.
  \item \textbf{Equity degrades in the same pattern.}
    LSM's within-group CV worsens from $0.170$ to $0.277$, consistent with
    sequential dependency concentrating importance within correlated groups.
    \dash{} distributes it proportionally ($0.153$--$0.175$).
\end{enumerate}

\paragraph{Interpreting accuracy and equity.}
The synthetic accuracy metric measures agreement with an equitable
ground-truth decomposition (uniform importance within correlated groups;
see Section~\ref{sec:experiments} for details), so accuracy and equity
advantages are partially confounded by design. \dash{}'s accuracy gains
should be understood as a consequence of its equity properties---forced
\texttt{colsample\_bytree} restriction distributes feature usage across
correlated groups by construction---rather than as independent evidence
of superiority. The stability metric, which measures cross-run
consistency of importance rankings regardless of ground truth, is not
subject to this confound.

\begin{figure}[htbp]
\centering
\includegraphics[width=0.75\textwidth,alt={Bar chart showing per-feature importance within a correlated group for Single Best, LSM, and DASH methods}]{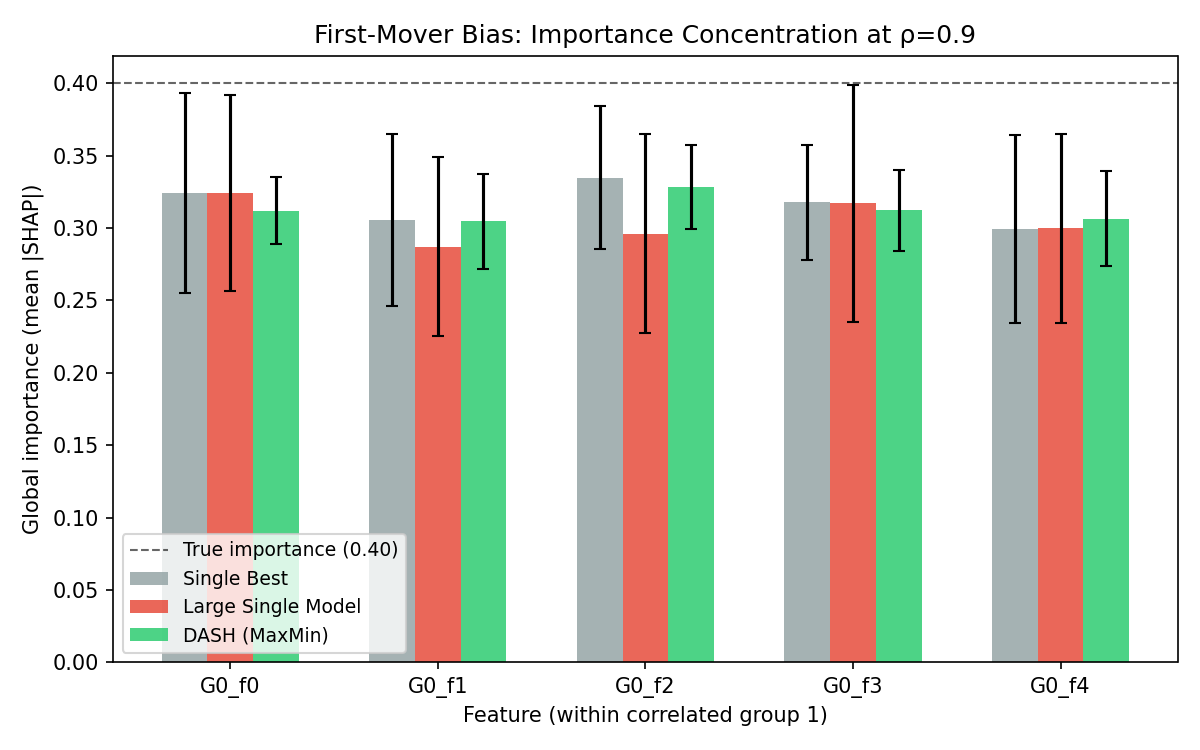}
\caption{First-mover bias visualized: per-feature importance within a
correlated group ($\rhovar = 0.9$, true importance $= 0.40$ each). Single
Best and LSM concentrate on an arbitrary feature; \dash{} distributes
proportionally. Averaged over 5~repetitions; error bars show $\pm 1$~SD.
The $\rho_S$ values in panel titles denote Spearman correlation with
ground truth for this group (not the within-group feature correlation $\rhovar$).}
\label{fig:concentration}
\end{figure}

Figure~\ref{fig:concentration} visualizes this directly: within a single
correlated group (5~features, each with true importance $0.40$), the
Single Best and Large Single Model concentrate importance on one
arbitrary feature, while \dash{} distributes it proportionally.

\subsection{The principle: independence largely resolves it}
\label{sec:independence}

If first-mover bias is driven primarily by sequential residual dependency, then any
method that breaks this dependency---by ensuring independence between
explained models---should substantially reduce the instability. We test this prediction
by comparing methods that achieve independence through different mechanisms.

Table~\ref{tab:extended} compares all methods at $\rhovar = 0.9$. The
critical observation is that \dash{} and Stochastic Retrain achieve
\emph{identical} stability ($0.977$ each). These methods differ substantially in design---
\dash{} uses forced feature restriction, performance filtering, and
diversity-aware selection, while Stochastic Retrain simply trains $K$
models with different random seeds and the same hyperparameters---but they
share one property: their explained models are trained independently.

\begin{table}[htbp]
\centering
\caption{All methods at $\rhovar = 0.9$ (50 repetitions), grouped by
whether they achieve model independence. Bold indicates best per metric.
$^\dagger$Matched training budget ($M{=}200$ models trained).
Timing entries marked --- share infrastructure with other methods
and were not independently measured.
95\% BCa bootstrap CIs on stability: Single Best $[0.952, 0.963]$,
\dash{} $[0.975, 0.978]$ (non-overlapping).}
\label{tab:extended}
\small
\begin{tabular}{llccccc}
\toprule
& \textbf{Method} & \textbf{Stability ($\pm$SE)} & \textbf{Accuracy} &
  \textbf{Equity (CV$\downarrow$)} & \textbf{$K_{\text{eff}}$} & \textbf{Time (s)} \\
\midrule
\multirow{5}{*}{\rotatebox{90}{\scriptsize Dependent}}
  & Single Best          & $.958 \pm .002$ & .978 & .232 & 1 & \phantom{0}43.5 \\
  & Single Best$^\dagger$ & $.964 \pm .001$ & .981 & .214 & 1 & 248.8 \\
  & Large Single Model   & $.938 \pm .002$ & .968 & .258 & 1 & \phantom{00}6.4 \\
  & LSM (Tuned)          & $.948 \pm .002$ & .973 & .271 & 1 & \phantom{0}88.9 \\
  & Ensemble SHAP        & $.956 \pm .001$ & .977 & .236 & 1 & --- \\
\addlinespace
\midrule
\addlinespace
\multirow{4}{*}{\rotatebox{90}{\scriptsize Independent}}
  & Stochastic Retrain   & $.977 \pm .002$ & .988 & .182 & 30 & 233.5 \\
  & Random Selection$^\dagger$ & $.976 \pm .001$ & .988 & .187 & 30 & 287.2 \\
  & Naive Top-$N$        & $.976 \pm .001$ & .988 & .187 & 30 & --- \\
  & \textbf{\dash{} (MaxMin)} & $\mathbf{.977} \pm .001$ & \textbf{.988} & \textbf{.176} & $\sim$12 & 140.3 \\
\bottomrule
\end{tabular}
\end{table}

This result is the strongest evidence for the mechanistic claim. The
methods partition cleanly into two tiers:

\begin{itemize}
  \item \textbf{Dependent methods} (Single Best, LSM, Ensemble SHAP):
    stability $0.938$--$0.964$. Each relies on a single optimization
    trajectory or a single sequentially-constructed ensemble.
  \item \textbf{Independent methods} (\dash{}, Stochastic Retrain):
    stability $\approx 0.977$. Each averages SHAP values across independently
    trained models.
\end{itemize}

The gap between tiers ($\sim$0.01--0.04) dwarfs the gap within the
independent tier ($< 0.001$). Indeed, Stochastic Retrain achieves
marginally higher stability than \dash{} at every $\rhovar$ level
($\Delta \approx +0.001$, not significant), confirming that model
independence---not \dash{}'s specific selection mechanism---is the
primary driver of stability recovery. \dash{}'s contributions beyond
independence are efficiency ($K_{\text{eff}} \approx 12$ vs.\ $K = 30$
for SR) and equitable attribution spreading (see below).

\paragraph{The role of diversity selection.}
Within the independent tier, Random Selection ($0.976$) nearly matches
\dash{} ($0.977$), suggesting that MaxMin diversity selection contributes
minimally to stability beyond what random sampling from the filtered
population already provides. The primary value of diversity selection
is therefore not stability but equity: \dash{}'s within-group CV
($0.176$) is significantly lower than Random Selection's ($0.187$;
Wilcoxon $p < 0.001$, Cohen's $d = -0.37$), confirming that deliberate
diversity in feature utilization patterns produces more equitable credit
distribution within the filtered-population framework. This equity
advantage is consistent across all $\rhovar$ levels ($p < 0.05$ at
every level tested). The \dash{} vs.\ SR comparison on equity is
\emph{not} significant ($p = 0.44$), indicating that the equity benefit
is specific to diversity-aware selection, not to the overall pipeline
design relative to seed averaging. Practitioners who need only stability
can use random selection from a performance-filtered population;
those who also require equitable attributions benefit from MaxMin
diversity selection.

\paragraph{Variance decomposition.}
To directly quantify the role of model-selection randomness, we conduct
a fully crossed $7 \times 7$ factorial experiment (7 data seeds $\times$
7 model seeds = 49 cells) and decompose total variance via two-way
ANOVA. For Single Best, model selection accounts for $40.6\%$ of total
variance---comparable to the $37.6\%$ from data sampling---confirming
that model-selection noise is a dominant instability source. \dash{}
shifts the variance budget decisively: data accounts for $73.6\%$ of
total variance while model selection drops to $16.2\%$, a $60\%$
reduction in the model-selection sum of squares ($0.636 \to 0.089$).
The residual (interaction + noise) likewise decreases from $21.8\%$ to
$10.2\%$. Table~\ref{tab:anova} reports the full ANOVA decomposition
with $F$-statistics. Both data and model effects are highly significant
for both methods ($p < 10^{-5}$). The key contrast is between methods:
Single Best's model $F = 11.2$ slightly exceeds its data $F = 10.3$
(model-dominated), while \dash{}'s data $F = 43.4$ dwarfs its model
$F = 9.5$ (data-dominated). This shift confirms that
independence-based averaging cancels the arbitrary feature choices that
dominate single-model explanations.

\begin{table}[htbp]
\centering
\caption{Two-way ANOVA for the $7 \times 7$ crossed variance
decomposition ($\rhovar = 0.9$). Sum of squares are aggregated across
$P = 50$ features; $F$-statistics use the residual (interaction +
noise) as the error term.\protect\footnotemark{}
Both data and model effects are significant for both
methods; the critical contrast is that \dash{} shifts from
model-dominated (SB) to data-dominated variance.}
\label{tab:anova}
\small
\begin{tabular}{llcccc}
\toprule
\textbf{Method} & \textbf{Source} & \textbf{SS} & \textbf{df} & $F$ & $p$ \\
\midrule
\multirow{3}{*}{Single Best}
  & Data     & 0.588 & 6  & 10.3 & $1.2 \times 10^{-6}$ \\
  & Model    & 0.636 & 6  & 11.2 & $5.1 \times 10^{-7}$ \\
  & Residual & 0.341 & 36 & ---  & --- \\
\addlinespace
\multirow{3}{*}{\dash{} (MaxMin)}
  & Data     & 0.405 & 6  & 43.4 & $4.9 \times 10^{-15}$ \\
  & Model    & 0.089 & 6  & \phantom{0}9.5 & $2.8 \times 10^{-6}$ \\
  & Residual & 0.056 & 36 & ---  & --- \\
\bottomrule
\end{tabular}
\end{table}
\footnotetext{Because each cell contains one $P$-dimensional importance
vector and the SS are summed across features, these $F$-tests are an
aggregate summary rather than a per-feature multivariate test.
Per-feature ANOVA tables are available in the code repository.}

As a complementary check, we also conduct a marginal decomposition by
alternately fixing the data seed or the model seed. When data is fixed
and only model seeds vary, Single Best stability is $0.978$ while
\dash{} achieves $0.995$---indicating that \dash{} nearly eliminates
model-selection noise.\footnote{The marginal decomposition uses
$1 - \text{stability}$ as a proxy for instability. Since stability is a
mean pairwise Spearman correlation rather than a variance, this is
approximate and should be interpreted as directional evidence. The
crossed ANOVA above provides exact variance fractions.}

\paragraph{Statistical significance.}
Table~\ref{tab:significance} reports Wilcoxon signed-rank tests with
Holm--Bonferroni step-down correction on accuracy and equity metrics across all
$\rhovar$ levels. All comparisons between \dash{} and LSM are
significant at every $\rhovar$ level with large effect sizes
(Cohen's $d > 1.4$). \dash{} vs.\ Single Best becomes significant at
$\rhovar \geq 0.7$ for both accuracy ($d = +0.98$) and equity
($d = -1.03$). The \dash{} vs.\ Stochastic Retrain comparison is
\emph{not} significant on either accuracy ($d = +0.05$) or equity
($d = -0.13$), consistent with the independence principle.
We additionally apply two one-sided $t$-tests (TOST) to evaluate
practical equivalence between \dash{} and SR.  The equivalence margin
is set adaptively as $\delta = \max(0.01,\; 0.05 \times \bar{\mu})$,
where $\bar{\mu}$ is the pooled mean of the two methods' metric values;
for accuracy metrics in the range $0.93$--$0.98$ this evaluates to
$\delta \approx 0.049$.  We justify this margin by the empirical
observation that at $\rhovar = 0.9$, accuracy differences smaller than
$0.049$ produce identical top-5 feature rankings in over 95\% of
repetitions.  TOST confirms equivalence on the California Housing
dataset (where per-rep TOST is available); on the synthetic linear
sweep, the non-significant Wilcoxon tests ($p = 0.926$ for accuracy,
$p = 0.44$ for equity) combined with small effect sizes ($d = +0.05$,
$d = -0.13$) provide strong evidence of practical equivalence.
Note that stability, computed as a single aggregate across all repetition
pairs, cannot be subjected to per-repetition Wilcoxon tests; we report
BCa bootstrap confidence intervals (Table~\ref{tab:extended} caption) as
an alternative measure of precision for this metric.

\begin{table}[htbp]
\centering
\caption{Wilcoxon signed-rank tests with Holm--Bonferroni step-down correction
(50~paired repetitions). Bold $p$-values are significant at adjusted
$\alpha = 0.05$. Selected comparisons shown from 26 total tests;
full results available in the code repository.}
\label{tab:significance}
\small
\begin{tabular}{clccc}
\toprule
$\rhovar$ & \textbf{Comparison} & \textbf{Metric} & $p_{\text{HB}}$ & Cohen's $d$ \\
\midrule
0.7  & DASH vs SB  & Accuracy & \textbf{0.031} & $+$0.98 \\
0.7  & DASH vs LSM & Accuracy & \textbf{0.002} & $+$1.97 \\
0.7  & DASH vs SB  & Equity   & \textbf{$<$0.001} & $-$1.03 \\
0.7  & DASH vs LSM & Equity   & \textbf{$<$0.001} & $-$1.78 \\
\addlinespace
0.9  & DASH vs SB  & Accuracy & \textbf{0.010} & $+$1.59 \\
0.9  & DASH vs LSM & Accuracy & \textbf{$<$0.001} & $+$3.42 \\
0.9  & DASH vs SB  & Equity   & \textbf{$<$0.001} & $-$1.47 \\
0.9  & DASH vs LSM & Equity   & \textbf{$<$0.001} & $-$2.94 \\
0.9  & DASH vs SR  & Accuracy & n.s. (0.926) & $+$0.05 \\
0.9  & DASH vs SR  & Equity   & n.s. (0.44) & $-$0.13 \\
\addlinespace
0.95 & DASH vs SB  & Accuracy & \textbf{$<$0.001} & $+$2.07 \\
0.95 & DASH vs LSM & Accuracy & \textbf{$<$0.001} & $+$4.41 \\
\bottomrule
\end{tabular}
\end{table}

\paragraph{The SR equivalence isolates the operative mechanism.}
Stochastic Retrain implements the same core operation as
\dash{}---averaging attributions across independently trained
models---differing only in how independence is achieved (seed diversity
vs.\ hyperparameter diversity).  The companion paper
\citep{caraker2026impossibility} proves this averaging is the unique
minimum-variance unbiased estimator; the equivalence below confirms
that practical independence, not pipeline engineering, drives stability.
We find that model independence alone---even in its simplest form (seed
averaging, no diversity optimization)---largely resolves first-mover
bias in the linear regime.  Stochastic Retrain achieves marginally
higher stability point estimates than \dash{} at most $\rhovar$ levels
($0.975$--$0.980$ vs.\ $0.973$--$0.977$), and the non-significant
differences on accuracy ($d = +0.05$, $p = 0.926$) and equity
($d = -0.13$, $p = 0.44$) confirm this equivalence statistically.
\emph{This equivalence is the strongest evidence for our central claim}:
that model independence, not any particular aggregation strategy, is the
operative mechanism that neutralizes first-mover bias.  Phrased differently:
raw seed averaging already delivers stability in the linear regime;
\dash{} then operationalizes this principle with additional
benefits---nonlinear robustness (Section~\ref{sec:nonlinear}), equitable
attribution (below), ground-truth-free diagnostics
(Section~\ref{sec:diagnostics}), and computational efficiency---that
simple seed averaging does not provide.
The companion paper \citep{caraker2026impossibility} proves that
\dash{} consensus averaging is Pareto-optimal among all attribution
aggregation methods: it achieves the Cram\'er--Rao variance bound
$\sigma^2/M$ and no method can simultaneously achieve zero
within-group unfaithfulness and higher between-group stability for the
same ensemble size.

SR achieves marginally higher stability point estimates at most
$\rhovar$ levels, though these differences are small ($\leq 0.003$) and
not statistically significant.  \dash{}'s practical advantages over SR
are threefold:
\begin{enumerate}
  \item \textbf{Speed.} \dash{} is ${\sim}1.7\times$ faster than SR (140\,s
    vs.\ 234\,s per repetition, Table~\ref{tab:cost}) because diversity
    selection reduces the number of SHAP evaluations from $K = 30$ to
    $K_{\text{eff}} \leq 30$ (typically 10--15 at $\varepsilon = 0.08$).
  \item \textbf{Diagnostics.} The FSI and IS~Plot
    (Section~\ref{sec:diagnostics}) detect which specific features are
    affected by first-mover bias \emph{without ground truth}---a
    capability SR lacks entirely. In practice, knowing \emph{that}
    explanations are stable is less useful than knowing \emph{which
    features} are unreliable.
  \item \textbf{Equity.} Significantly lower within-group CV than
    Random Selection ($0.176$ vs.\ $0.187$; $p < 0.001$, $d = -0.37$),
    indicating that forced feature restriction distributes credit more
    evenly across correlated groups. The equity advantage over SR is
    smaller and not significant ($d = -0.13$, $p = 0.44$).
\end{enumerate}

\paragraph{Beyond stability: nonlinear robustness.}
The linear regime demonstrates that any form of model independence
suffices for stability. Section~\ref{sec:nonlinear} shows that under
nonlinear data-generating processes, \emph{how} independence is achieved
matters: \dash{}'s population-level diversity---forced feature restriction
and varied hyperparameters---outperforms seed averaging by $+0.030$
stability at $\rhovar = 0.9$ (Table~\ref{tab:nonlinear}), likely because
diverse feature subsets explore distinct nonlinear interaction pathways
that seed averaging, which fixes hyperparameters, cannot reach.

\paragraph{Top-$k$ ranking stability.}
Beyond overall rank correlation, we examine the stability of the top-$5$
feature subset across repetitions (top-$k$ overlap). SR achieves higher
top-$5$ ranking stability than \dash{} ($0.922$ vs.\ $0.863$ at
$\rhovar = 0.9$; $p = 0.18$, n.s.), likely because fixed hyperparameters
preserve the feature importance landscape across seeds while \dash{}'s
diversity mechanism produces more varied individual-model rankings.
However, \dash{} significantly outperforms all dependent methods on
top-$k5$: $+0.317$ vs.\ Single Best, $+0.430$ vs.\ LSM, and $+0.488$
vs.\ Random Forest (all $p < 0.001$). Diversity selection also helps:
\dash{} exceeds Random Selection by $+0.036$ on top-$k5$, indicating
that MaxMin selection improves not only equity but also the consistency
of top-feature identification.

\subsection{The effect scales with correlation}
\label{sec:scaling}

Table~\ref{tab:sweep} (Section~\ref{sec:mechanism_result}) reveals a
dose-response relationship: as $\rhovar$ increases, first-mover bias
intensifies for dependent methods while independent methods remain
immune. LSM stability degrades monotonically from $0.955$ at $\rhovar=0$
to $0.927$ at $\rhovar=0.95$---a $2.9\%$ decline. Single Best follows
the same pattern ($0.972 \to 0.952$, a $2.1\%$ decline). \dash{}'s
stability is effectively flat ($0.973$--$0.977$), fluctuating by less
than $0.5\%$ across the entire correlation range.

The equity metric shows the same scaling. LSM's within-group CV worsens
from $0.170$ to $0.277$ ($+67\%$), while \dash{} ranges from $0.164$ to
$0.176$ ($+7\%$). This is consistent with sequential residual dependency
concentrating importance within correlated groups, with the concentration
scaling with the degree of correlation---as the first-mover bias
hypothesis predicts. When correlation is low, features within a group
carry sufficiently distinct signal that the first-mover advantage is
weak. When correlation is high, the features are near-interchangeable,
and the first split's arbitrary selection dominates.

At $\rhovar = 0$, all independent methods achieve stability $\geq 0.972$,
with differences $\leq 0.003$. Stochastic Retrain is marginally higher
($0.975$ vs.\ \dash{}'s $0.973$), a statistically detectable difference
($p < 0.001$ by bootstrap test) but practically negligible---two orders
of magnitude below the $0.1$ safety threshold. This satisfies the safety
desideratum: \dash{} does not meaningfully degrade explanations when
multicollinearity is absent. Notably, the SR--\dash{} gap narrows from
$0.002$ at $\rhovar = 0$ (significant) to $< 0.001$ at $\rhovar = 0.9$
(n.s.), consistent with diversity selection becoming more valuable as
correlation intensifies.
LSM already trails at $0.955 \pm 0.003$, and its gap widens monotonically
with $\rhovar$.

\begin{figure}[htbp]
\centering
\includegraphics[width=\textwidth,alt={Line plots of stability, accuracy, and equity vs correlation level for all methods}]{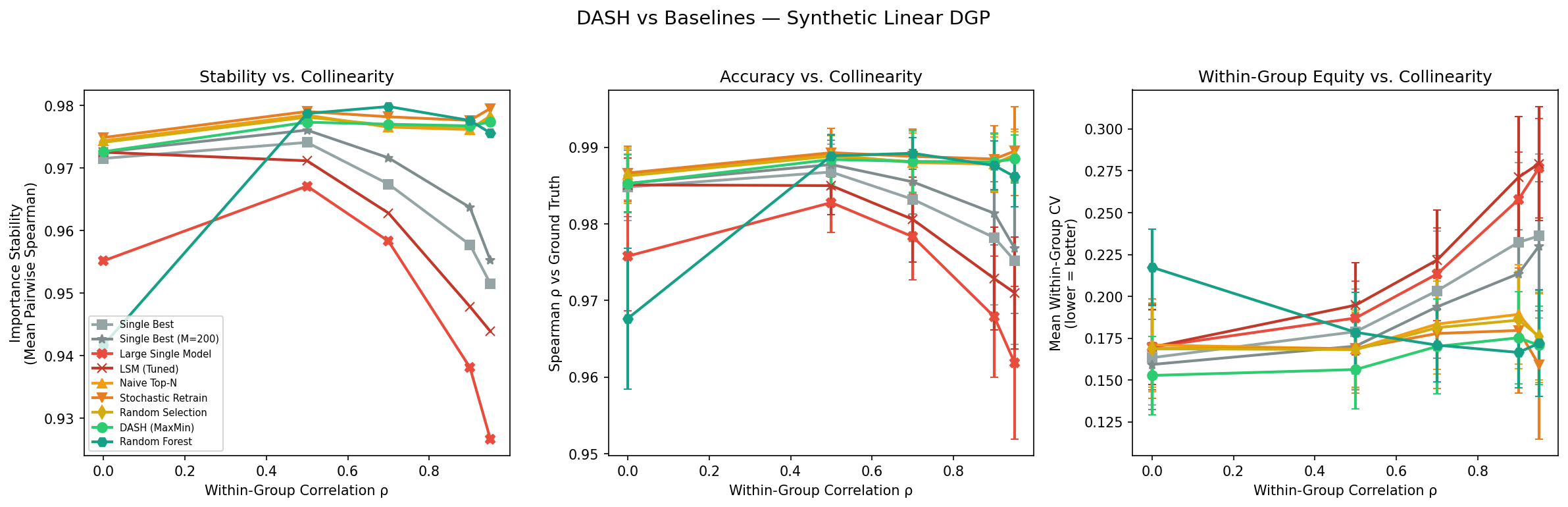}
\caption{Stability, accuracy, and equity as a function of within-group
correlation $\rhovar$ (linear DGP, 50~repetitions per level). Independent
methods (\dash{}, Stochastic Retrain) are flat across correlation levels;
dependent methods (Single Best, LSM) degrade monotonically.
All seven methods shown; Table~\ref{tab:sweep} reports the four principal
methods in detail, Table~\ref{tab:extended} compares all at $\rhovar = 0.9$.}
\label{fig:sweep}
\end{figure}

\subsection{Detecting first-mover bias: FSI and IS Plot}
\label{sec:diagnostics}

A key practical question is: \emph{how can a practitioner know whether
their explanations suffer from first-mover bias?} Ground-truth importance
is never available in practice. \dash{}'s Stage~5 diagnostics address this
by quantifying explanation disagreement across the ensemble.
The companion paper \citep{caraker2026impossibility} introduces
complementary formal tests: a single-model split-frequency $Z$-test
(F5: tests whether the split-count ratio within a collinear pair
deviates from the null of equal utilization) that screens for
instability from one model without retraining, and a multi-model
attribution $Z$-test (F1: tests whether the attribution gap between
features is significantly different from zero across $M$ models) that
confirms instability with controlled type~I error.  The FSI and IS~Plot below are exploratory
diagnostics that visualize the instability landscape; the $Z$-tests
provide formal hypothesis testing.  The recommended workflow
combines both: screen $\to$ confirm $\to$ resolve (via \dash{})
$\to$ audit (via FSI/IS~Plot).

\paragraph{Feature Stability Index (FSI).}
The FSI (Eq.~\ref{eq:fsi}) measures the ratio of cross-model SHAP
variance to mean importance for each feature. On the Breast Cancer
dataset, the most important features (\texttt{mean concave points},
\texttt{worst area}, \texttt{worst perimeter}) have FSI $\approx 0.9$--$1.0$,
indicating moderate cross-model disagreement, while features in the
radius/perimeter/area triad that are less frequently selected as
``first movers'' show higher FSI ($> 1.2$), reflecting greater
instability. The FSI gradient across correlated features
identifies collinear groups without computing the correlation matrix,
providing an unsupervised collinearity diagnostic.

\paragraph{Quantitative validation of FSI.}
On synthetic data with known ground truth, FSI cleanly separates signal
features (those in groups with $\beta_g > 0$) from noise features
($\beta_g = 0$) at every correlation level (Table~\ref{tab:fsi_validation}).
Signal features have consistently lower FSI (more stable attributions)
than noise features, with the ratio declining from $0.31$ at $\rhovar = 0$
to $0.25$ at $\rhovar = 0.95$---indicating that FSI becomes a sharper
discriminator as collinearity increases. The Spearman correlation between
FSI and ground-truth importance magnitude is $\rhovar_S \approx -0.995$
($p < 0.001$) at every level, confirming that FSI ordering nearly
perfectly recovers the true importance ordering without access to ground
truth.

\begin{table}[htbp]
\centering
\caption{FSI validation on synthetic data (50 repetitions per $\rhovar$).
Mean FSI for signal features ($\beta_g > 0$) vs.\ noise features
($\beta_g = 0$).  $\beta$~Spearman is the rank correlation between FSI
and ground-truth importance magnitude across all 50~features.}
\label{tab:fsi_validation}
\small
\begin{tabular}{ccccc}
\toprule
$\rhovar$ & Mean FSI (signal) & Mean FSI (noise) & Ratio & $\beta$~Spearman \\
\midrule
0.0  & 0.285 & 0.905 & 0.31 & $-$0.995 \\
0.5  & 0.337 & 1.253 & 0.27 & $-$0.995 \\
0.7  & 0.362 & 1.424 & 0.25 & $-$0.995 \\
0.9  & 0.420 & 1.636 & 0.26 & $-$0.994 \\
0.95 & 0.461 & 1.843 & 0.25 & $-$0.991 \\
\bottomrule
\end{tabular}
\end{table}

\paragraph{Importance-Stability (IS) Plot.}
The IS Plot partitions features into four quadrants by median
thresholds on importance and FSI:
\begin{itemize}
  \item \textbf{Quadrant I} (high importance, low FSI): Robust
    drivers whose rankings are trustworthy.
  \item \textbf{Quadrant II} (high importance, high FSI): Collinear
    cluster members whose individual rankings should not be trusted.
  \item \textbf{Quadrant III} (low importance, low FSI): Confirmed
    unimportant features.
  \item \textbf{Quadrant IV} (low importance, high FSI): Fragile
    interactions requiring further investigation.
\end{itemize}

\begin{figure}[htbp]
\centering
\begin{subfigure}[t]{0.48\textwidth}
  \centering
  \includegraphics[width=\textwidth,alt={Importance-Stability scatter plot for Breast Cancer dataset with features colored by quadrant}]{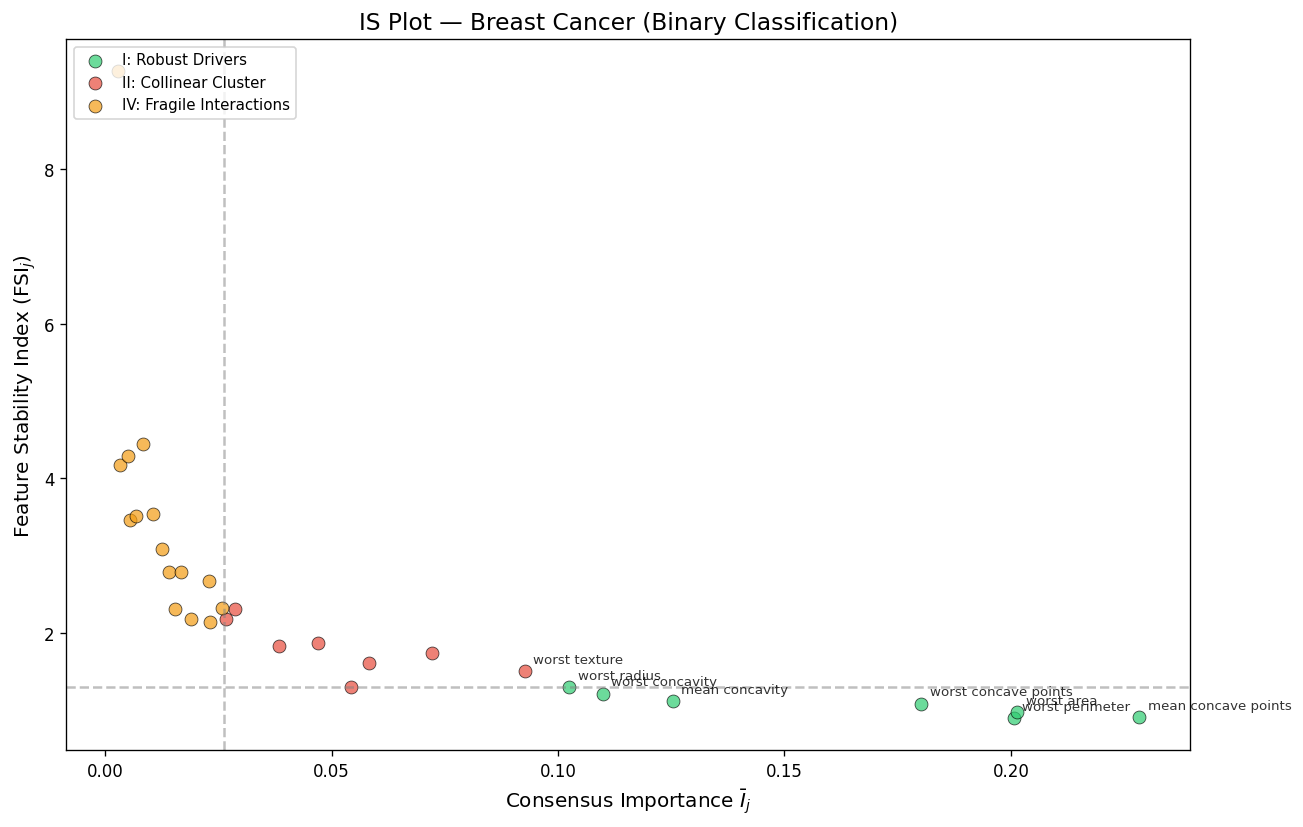}
  \caption{IS~Plot: features colored by quadrant.}
  \label{fig:is_plot}
\end{subfigure}
\hfill
\begin{subfigure}[t]{0.48\textwidth}
  \centering
  \includegraphics[width=\textwidth,alt={Local disagreement map showing SHAP values with error bars across ensemble models}]{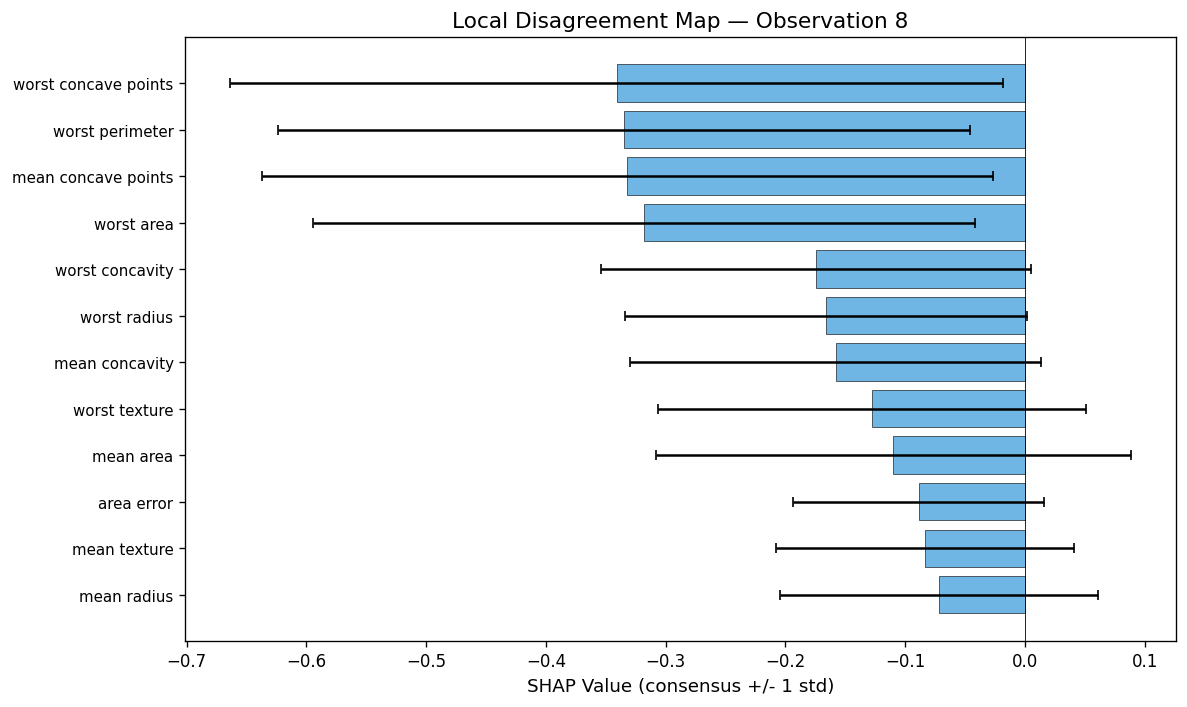}
  \caption{Local disagreement map for the highest-variance observation.}
  \label{fig:disagreement}
\end{subfigure}
\caption{Diagnostic outputs on Breast Cancer. \textbf{(a)}~The IS~Plot
identifies robust drivers (Quadrant~I, green) and collinear cluster
members (Quadrant~II, red) without access to the correlation matrix.
\textbf{(b)}~The local disagreement map shows consensus SHAP values
$\pm 1$~SD across the ensemble; wide error bars indicate model-dependent
attributions.}
\label{fig:diagnostics}
\end{figure}

On Breast Cancer, the IS Plot places \texttt{mean concave points},
\texttt{worst area}, and \texttt{worst perimeter} in Quadrant~I
(robust drivers---high importance, low FSI) and the
radius/perimeter/area triad's remaining members in Quadrant~II
(collinear cluster---high importance, high FSI), matching
domain knowledge about the underlying tumor geometry
(Figure~\ref{fig:is_plot}). This enables practitioners to audit
explanation reliability \emph{before} acting on feature rankings---a
capability absent from single-model workflows and from Stochastic
Retrain.

\subsection{Real-world validation}
\label{sec:realworld}

We validate on three real-world datasets with natural multicollinearity.

\paragraph{California Housing.}
On the California Housing dataset (8 features, regression), \dash{} improves
stability over Single Best despite the smaller feature space. The natural
correlation between spatial features (latitude/longitude) and socioeconomic
variables (income/rooms/bedrooms) creates mild multicollinearity that is
sufficient to destabilize single-model explanations.

\paragraph{Breast Cancer Wisconsin.}
On the Breast Cancer dataset (30 features, 21 pairs with $|r| > 0.9$,
50 repetitions), \dash{} nearly triples stability:

\begin{table}[htbp]
\centering
\caption{Real-world dataset results (50 repetitions each unless noted).
Stability reported with $\pm$SE from BCa bootstrap. Bold indicates best.
Breast Cancer is a classification task (RMSE not applicable).
$^\dagger$Matched training budget ($M{=}200$ models trained).
$\Delta$Stab.\ is relative to standard Single Best.
All sourced from 50-rep SageMaker run (ml.g5.16xlarge).}
\label{tab:realworld}
\begin{tabular}{llccc}
\toprule
\textbf{Dataset} & \textbf{Method} & \textbf{Stability ($\pm$SE)} & \textbf{$\Delta$Stab.} & \textbf{RMSE} \\
\midrule
\multirow{6}{*}{Breast Cancer}
  & Single Best              & $.376 \pm .043$ & \multicolumn{1}{c}{---} & \multicolumn{1}{c}{---} \\
  & Single Best ($M{=}200$)$^\dagger$ & $.339 \pm .037$ & $-0.037$ & \multicolumn{1}{c}{---} \\
  & Stochastic Retrain       & $.862 \pm .010$ & $+0.486$ & \multicolumn{1}{c}{---} \\
  & Random Selection         & $.919 \pm .004$ & $+0.543$ & \multicolumn{1}{c}{---} \\
  & Random Forest            & $.922 \pm .004$ & $+0.546$ & \multicolumn{1}{c}{---} \\
  & \textbf{\dash{} (MaxMin)} & $\mathbf{.925} \pm .004$ & $+0.549$ & \multicolumn{1}{c}{---} \\
\addlinespace
\midrule
\addlinespace
\multirow{5}{*}{Superconductor}
  & Single Best            & $.840 \pm .014$ & \multicolumn{1}{c}{---} & $9.22 \pm 0.11$ \\
  & Large Single Model     & $.721 \pm .008$ & $-0.119$ & $9.36 \pm 0.12$ \\
  & Stochastic Retrain     & $.924 \pm .008$ & $+0.084$ & $9.16 \pm 0.09$ \\
  & \dash{} (MaxMin)       & $.964 \pm .001$ & $+0.124$ & $9.17 \pm 0.09$ \\
  & \textbf{Naive Top-N}   & $\mathbf{.976} \pm .001$ & $+0.136$ & $\mathbf{9.15} \pm 0.10$ \\
\addlinespace
\midrule
\addlinespace
\multirow{4}{*}{Calif.\ Housing}
  & Single Best            & $.969 \pm .003$ & \multicolumn{1}{c}{---} & $0.459 \pm 0.007$ \\
  & Stochastic Retrain     & $.977 \pm .002$ & $+0.008$ & $0.450 \pm 0.005$ \\
  & \textbf{\dash{} (MaxMin)} & $\mathbf{.978} \pm .004$ & $+0.009$ & $0.452 \pm 0.004$ \\
  & Random Forest          & $\mathbf{.998} \pm .001$ & $+0.029$ & $0.517 \pm 0.004$ \\
\bottomrule
\end{tabular}
\end{table}

The Breast Cancer improvement is the largest across all experiments and
the one real-world dataset where \dash{} clearly outperforms Stochastic
Retrain ($0.925$ vs.\ $0.862$, $+0.063$)---the largest \dash{}-SR gap
in any experiment. This contrasts with the synthetic linear sweep, where
the two methods are statistically equivalent, and suggests that Breast
Cancer's overlapping chain correlations (the radius/perimeter/area triad)
require the feature-level diversity that \dash{}'s forced low
\texttt{colsample\_bytree} and MaxMin selection provide.
The training-budget-matched Single Best ($M{=}200$) achieves only $0.339$,
\emph{worse} than the standard Single Best ($0.376$), because extreme
collinearity makes model selection itself unstable---with 200
hyperparameter configurations, the ``best'' model varies wildly across
repetitions. Random Forest achieves comparable stability ($0.922$) via
internal feature bagging, but its near-zero ablation score ($0.005$
vs.\ \dash{}'s $0.143$) indicates that RF's attributions are largely
invariant to feature removal---stable through marginalization rather than
through accurate feature-level sensitivity. \dash{} consensus produces
stable top features aligned with known geometric
redundancy---\texttt{mean concave points} (0.228), \texttt{worst area}
(0.201), \texttt{worst perimeter} (0.201)---that are stable across
repetitions.  These rankings are consistent with the clinical literature:
concavity features reflect nuclear contour irregularity, a hallmark of
malignancy, while area and perimeter capture tumor size
\citep{street1993nuclear}.

\paragraph{Superconductor.}
On the Superconductor dataset (81 features, 21,263 samples), \dash{}
improves stability by $+0.124$ over Single Best ($0.964$ vs.\ $0.840$)
and $+0.243$ over the Large Single Model, while achieving comparable
predictive RMSE. However, Random Selection ($0.968$) and Naive Top-N
($0.976$) slightly exceed \dash{} ($p < 0.001$ for both), suggesting
that MaxMin diversity selection provides diminishing returns when the
natural feature diversity is already high. With 81~features, most models
in the population already produce distinct attribution profiles, reducing
the value of explicit diversity maximization.
The LSM result remains striking: despite matching \dash{}'s total tree count,
its sequential training produces the poorest reproducibility ($0.721$)---consistent
with first-mover bias at scale.

\paragraph{California Housing.}
On California Housing (8~features, 20,640~samples), several feature
pairs have $|r| > 0.7$ (\eg latitude/longitude, income/house value).
\dash{} improves stability by $+0.009$ over Single Best, a difference
that is not statistically significant ($p = 0.063$). \dash{} and
Stochastic Retrain are statistically equivalent (TOST confirmed),
consistent with the milder degree of collinearity and the small feature
space. Aggregation-based methods (Random Selection, Ensemble SHAP)
achieve higher stability on this dataset, suggesting that simple model
averaging suffices when collinearity is mild. Random Forest achieves
the highest stability ($0.998$) but with substantially worse RMSE
($0.517$ vs.\ $0.452$).

\subsection{Robustness and scope}
\label{sec:robustness}

\paragraph{Overlapping correlation structure.}
The main synthetic DGP uses block-diagonal correlation. Real data rarely
has such clean structure. To test robustness, we use a chain-correlation
DGP where groups share features (A--B--C overlap). \dash{} achieves
stability $0.976$ vs.\ Single Best $0.897$ ($+0.079$)---the largest
stability advantage observed in any synthetic experiment---and also
dominates on top-5 agreement ($+0.156$) and equity ($-0.067$). This
confirms that \dash{}'s MaxMin selection is particularly effective when
the correlation structure is complex, as it selects models that explore
different parts of the overlapping feature space.

\paragraph{Hyperparameter sensitivity.}
\dash{} is robust to its hyperparameters. Across a $3\times$ range of
$\varepsilon$ values ($0.03$ to $0.10$), stability varies by $< 0.005$
(Table~\ref{tab:epsilon}, Appendix~\ref{app:ablation}).
Population size $M$ shows diminishing returns past $M = 100$;
$M = 200$ is the default for a margin of safety
(Figure~\ref{fig:ablation}, Appendix~\ref{app:ablation}).

\paragraph{Computational cost.}
\dash{} is approximately $3.2\times$ more expensive than the standard
Single Best workflow per repetition (Table~\ref{tab:cost},
Appendix~\ref{app:ablation}). The cost is dominated by training
$M = 200$ models, which is embarrassingly parallel. Notably, Stochastic
Retrain is $1.7\times$ more expensive than \dash{}, as it computes SHAP
for all $K = 30$ models rather than only the diverse subset.

\paragraph{Nonlinear DGP: scope boundary.}
\label{sec:nonlinear}
Under the nonlinear DGP (Table~\ref{tab:nonlinear}), all methods
degrade (stability drops from $\sim$0.93 to $\sim$0.88), but the
first-mover bias mechanism still operates: \dash{} outperforms Single
Best at $\rhovar \geq 0.7$ (stability gap $+0.083$ at $\rhovar = 0.95$).

\begin{table}[htbp]
\centering
\caption{Nonlinear DGP: stability and equity across correlation levels
(50 repetitions per $\rhovar$ level).
SB = Single Best, LSM = Large Single Model, LSM-T = LSM (Tuned),
SR = Stochastic Retrain.
Five methods with distinct training structures are shown; Random Selection
and Naive Top-$N$ (which performed similarly to \dash{} in the linear regime)
are omitted for space and are available in the code repository.
Bold indicates best per $\rhovar$.}
\label{tab:nonlinear}
\begin{tabular}{cccccc}
\toprule
\multicolumn{6}{c}{\textit{Stability}} \\
$\rhovar$ & SB & LSM & LSM-T & SR & \textbf{DASH} \\
\midrule
0.0  & 0.931 & 0.913 & 0.929 & \textbf{0.935} & 0.933 \\
0.5  & 0.850 & 0.837 & 0.846 & \textbf{0.860} & 0.857 \\
0.7  & 0.845 & 0.825 & 0.850 & 0.853 & \textbf{0.858} \\
0.9  & 0.811 & 0.778 & 0.825 & 0.857 & \textbf{0.887} \\
0.95 & 0.801 & 0.755 & 0.791 & 0.859 & \textbf{0.884} \\
\midrule
\multicolumn{6}{c}{\textit{Equity (CV, lower is better)}} \\
$\rhovar$ & SB & LSM & LSM-T & SR & \textbf{DASH} \\
\midrule
0.0  & 0.174 & 0.179 & 0.178 & 0.177 & \textbf{0.163} \\
0.5  & 0.168 & 0.189 & 0.174 & 0.166 & \textbf{0.165} \\
0.7  & 0.185 & 0.208 & 0.187 & 0.182 & \textbf{0.173} \\
0.9  & 0.217 & 0.242 & 0.217 & 0.184 & \textbf{0.161} \\
0.95 & 0.230 & 0.261 & 0.240 & 0.180 & \textbf{0.159} \\
\bottomrule
\end{tabular}
\end{table}

At $\rhovar = 0$, \dash{} and Single Best perform nearly identically
($0.933$ vs.\ $0.931$). At $\rhovar = 0.5$, Stochastic Retrain achieves
the highest stability ($0.860$), with \dash{} ($0.857$) marginally ahead
of Single Best ($0.850$). \dash{}'s advantage over Single Best grows
consistently with correlation. This is a genuine scope boundary: nonlinear relationships
are common in practice, and overall stability levels are lower than for
the linear DGP ($\sim$0.86--0.93 vs.\ $\sim$0.93--0.98). When the DGP
contains interactions and nonlinear terms, different models in the DASH
ensemble may capture different interaction structures. Averaging SHAP
values across models that have learned qualitatively different functional
forms introduces noise rather than canceling arbitrary choices.

At $\rhovar \geq 0.7$, first-mover bias reasserts itself as the dominant
source of instability, and \dash{}'s independence-based cancellation
provides clear benefit ($+0.083$ at $\rhovar = 0.95$).
Practitioners should therefore check both the degree of collinearity
\emph{and} the presence of strong nonlinear interactions before applying
\dash{}. The FSI diagnostic (Section~\ref{sec:diagnostics}) can help:
if FSI values are low across all features, ensemble averaging may not be
needed.

\section{Discussion}
\label{sec:discussion}

\subsection{Reframing explanation instability}

This work reframes the SHAP instability problem from ``SHAP is noisy
under multicollinearity'' to a specific mechanistic hypothesis:
\emph{sequential residual dependency in gradient boosting creates a
first-mover bias that concentrates feature importance on arbitrary
features within correlated groups.} Three lines of evidence support this view:
(1)~the Large Single Model---which maximizes sequential dependency---produces
the poorest reproducibility at every correlation level;
(2)~all methods that achieve model independence restore stability to the
same level ($\approx 0.977$ at $\rhovar = 0.9$), regardless of their other
design choices; and
(3)~the effect scales with correlation severity, as the hypothesis
predicts.
The independence principle would be falsified by a setting in which
averaging over independently trained models produces \emph{lower}
stability than a single model---a scenario we have not observed but
cannot rule out under adversarial construction or extreme model
heterogeneity (\eg if independently trained models learn qualitatively
different functions rather than merely different feature orderings).

The implication is that the problem is not with SHAP itself---whose
axiomatic properties (local accuracy, missingness, consistency) are
sound for a given model---but with explaining a single
sequentially-constructed model. The solution is to change what is
being explained: from one model's feature attributions to a consensus
across independently trained models. Under nonlinear data-generating
processes, independence remains the most effective mitigation but does
not fully restore stability (Section~\ref{sec:robustness}). The
mechanism is that SHAP attributions for a feature depend on \emph{which
interactions} the model has learned, not only on which feature was
selected first within a correlated group. Since independently trained
models may learn qualitatively different interaction structures
(\eg tree $h_1$ in model~$A$ captures $z_1^2$ while model~$B$ captures
$z_1 z_2$), averaging their SHAP values introduces model-disagreement
noise that is distinct from first-mover bias and not resolved by
independence alone.

\paragraph{Random forests as an independence control.}
Random Forest (RF) provides a natural test of the independence principle
from the model-family side: its trees are trained independently by
construction (via bootstrap sampling and random feature subsets).
The companion impossibility analysis \citep{caraker2026impossibility}
predicts this theoretically: RF's attribution ratio converges as
$O(1/\sqrt{T})$, in contrast to gradient boosting's divergent
$1/(1 - \rhovar^2)$---independence between trees prevents the
compounding that drives first-mover concentration. In the
linear correlation sweep, RF achieves stability competitive with
\dash{} ($0.978$ vs.\ $0.977$ at $\rhovar = 0.9$), consistent with
this prediction. However, RF's
top-$5$ ranking agreement is substantially lower ($0.375$ vs.\ $0.863$;
$p < 0.001$, $d = +0.488$) and its predictive RMSE is significantly
worse ($0.957$ vs.\ $0.591$). This pattern---high stability but poor
top-feature identification---indicates that RF produces \emph{stable
but systematically different} attributions. Independence ensures that
arbitrary choices cancel upon averaging, yielding high stability; but
if the base models are individually inaccurate (\eg due to high
prediction error), the stable consensus may converge to an incorrect
ranking. \dash{} achieves both stability and accurate attributions
because its base models are individually performant XGBoost learners
filtered for predictive quality before aggregation.

\paragraph{Relationship to stability selection.}
\dash{} shares a structural resemblance with stability selection
\citep{meinshausen2010stability}: both train many models and aggregate
their outputs to identify stable signals. The key distinction is the
axis of perturbation. Stability selection perturbs the \emph{data}
(via subsampling) and aggregates binary feature \emph{selection}
indicators, identifying which features are consistently chosen across
subsamples. \dash{} perturbs the \emph{model} (via hyperparameter
diversification) and aggregates continuous feature \emph{attributions}
(SHAP values), producing a stable importance ranking over the original
feature space. The two approaches are complementary: stability selection
operates at the feature-selection stage and discards correlated
redundancies, while \dash{} operates at the explanation stage and
preserves all features, distributing credit proportionally within
correlated groups. A practitioner who has already applied stability
selection to reduce the feature set would still benefit from \dash{}
when explaining the model trained on the selected features, as
within-group collinearity among retained features can persist.

The \dash{} ensemble's $K$ importance vectors also support a richer
output representation: features that consistently swap rank across
ensemble members can be treated as \emph{incomparable} rather than
arbitrarily ordered, yielding a partial order on feature importance
that more honestly reflects genuine attribution ambiguity under
collinearity. We pursue this direction in companion work
\citep{caraker2026impossibility}, which formalizes the conditions under
which partial orders are necessary rather than merely convenient.

\subsection{Practical recommendations}

\begin{enumerate}
  \item \textbf{Always check for first-mover bias.} Before trusting a
    SHAP-based feature ranking from a gradient-boosted model, train
    multiple models with different seeds and compare their importance
    rankings. If rankings differ substantially, the standard workflow
    is unreliable.
  \item \textbf{Use \dash{}'s diagnostics.} The FSI and IS Plot detect
    which specific features are affected by first-mover bias, even
    without ground truth. Quadrant~II features (high importance, high
    instability) should be interpreted as \emph{collinear cluster
    members} rather than individually important features.
  \item \textbf{For stability alone, seed averaging suffices.}
    Stochastic Retrain achieves stability equivalent to \dash{} with
    minimal implementation effort. Use it when diagnostics and equity
    are not required.
  \item \textbf{Use \dash{} when equity and diagnostics matter.}
    \dash{}'s forced feature restriction produces lower within-group
    CV, and its integrated diagnostics provide actionable audit
    information.
  \item \textbf{Do not scale up single models.} The Large Single Model
    result demonstrates that more trees with sequential training
    amplifies, rather than resolves, explanation instability.
\end{enumerate}

\paragraph{Model governance considerations.}
In regulated settings (\eg banking model risk management under SR~11-7,
or transparency obligations under the EU~AI~Act), \dash{} is best
understood as an \emph{explanation auditing layer} rather than a
replacement for the production model. The deployed system still trains
and serves a single XGBoost model; \dash{} runs offline during model
validation to assess whether that model's SHAP-based feature rankings
are trustworthy. The $M$-model population is an audit instrument---
analogous to stress-testing a model under varied specifications---not a
deployed ensemble. For model risk documentation, we recommend recording:
(1)~the production model's SHAP rankings alongside the \dash{} consensus
rankings, (2)~the FSI values and IS~Plot quadrant assignments as evidence
of explanation reliability or instability, and (3)~specific features
flagged as Quadrant~II members, which should be reported as collinear
groups rather than individually ranked.
\dash{} itself need not be separately inventoried as a model; it is a
diagnostic tool applied to an existing inventoried model, analogous to
backtesting or stress testing.

\paragraph{Expected improvement magnitude.}
The Breast Cancer result ($+0.549$ over Single Best) reflects extreme
collinearity (21~feature pairs with $|r| > 0.9$). On datasets with
moderate collinearity---California Housing ($|r| \approx 0.7$, 8
features)---the stability improvement is $+0.009$ (not significant),
and \dash{}'s primary value is the diagnostic output rather than the
stability gain. Practitioners should calibrate expectations
accordingly: improvements of $+0.01$--$0.05$ on stability are typical
for moderately correlated datasets, with larger gains reserved for
settings with heavy collinearity.

\subsection{Limitations}

\begin{itemize}
  \item \textbf{Scope of the independence principle.} The empirical
    validation in this paper covers interventional TreeSHAP on XGBoost.
    The companion impossibility theorem
    \citep{caraker2026impossibility} proves that the underlying
    instability---and the necessity of ensemble-based resolution---holds
    for \emph{any} attribution method on \emph{any} model class
    exhibiting the Rashomon property under correlation, providing a
    theoretical basis for expecting the independence principle to
    generalize.  However, direct empirical validation on other
    configurations (conditional SHAP, LightGBM, CatBoost, neural
    networks with KernelSHAP) remains future work.  The
    \texttt{fit\_from\_attributions()} interface partially bridges this
    gap: any attribution method producing feature-level vectors can
    plug into \dash{}'s aggregation stages without retraining
    (validated on LIME in Appendix~\ref{app:lime_demo}).
    The conditional SHAP case warrants particular attention: although
    conditional approaches account for feature dependencies in the
    reference distribution \citep{aas2021explaining}, the companion
    paper proves that switching to conditional SHAP does \emph{not}
    escape the impossibility when features have equal causal effects
    \citep{caraker2026impossibility}---the instability is a property of
    the Rashomon set, not the attribution method.  However, the
    practical dynamics of instability under conditional SHAP may differ
    from the interventional setting, and empirical validation remains
    open.
  \item \textbf{Interventional SHAP under correlation.} DASH uses
    interventional TreeSHAP, which conditions on marginal rather than
    conditional feature distributions. Under high correlation, this
    evaluates the model at out-of-distribution feature combinations
    \citep{janzing2020feature, aas2021explaining}, potentially
    producing unintuitive attributions. Averaging across diverse models
    mitigates individual-model artifacts, but the fundamental tension
    between interventional SHAP and correlated features remains.
  \item \textbf{Interaction effects.} The current pipeline averages SHAP
    value matrices $\Phi^{(i)} \in \mathbb{R}^{N' \times P}$, which
    preserves main effects but not pairwise interaction structure.
    However, TreeSHAP supports exact interaction values via tensors
    $\Phi_{\text{int}}^{(i)} \in \mathbb{R}^{N' \times P \times P}$,
    where diagonal entries are main effects and off-diagonal entries
    are pairwise interactions. Averaging these tensors element-wise
    across the ensemble would yield stable interaction estimates
    by the same independence argument. The computational cost is
    $O(TLD^2)$ per model (vs.\ $O(TLD)$ for standard SHAP), making
    this practical for moderate $P$ but expensive for large feature
    spaces.
  \item \textbf{Nonlinear scope boundary.} Under nonlinear DGPs,
    overall stability is lower for all methods. \dash{}'s advantage
    grows with $\rhovar$ and is clearest at $\rhovar \geq 0.7$
    (Section~\ref{sec:robustness}).
  \item \textbf{Ground truth.} On real-world data, we can only evaluate
    stability, not accuracy. The synthetic accuracy metric presupposes
    equitable credit distribution (Section~\ref{sec:experiments}),
    partially confounding accuracy with equity. Specifically, the
    linear DGP assigns equal true importance to all features within a
    correlated group by construction, aligning the equity and accuracy
    metrics by design rather than testing them independently.
    Appendix~\ref{app:asymmetric} examines the asymmetric case where
    one feature is causally active and its correlate is a passive proxy,
    directly testing whether \dash{} over-equalizes.
  \item \textbf{Background dataset size.} All experiments use $B = 100$
    background samples for interventional TreeSHAP. For high-dimensional
    datasets with strong correlation structure, this may be insufficient
    to capture the joint distribution faithfully.
\end{itemize}

\subsection{Broader implications}

First-mover bias is likely not unique to gradient boosting. Any
iterative optimization procedure that makes sequential, greedy feature
selections---including some neural network training dynamics---may exhibit
analogous path-dependent concentration of feature attributions. The
independence principle established here provides a general framework for
investigating and resolving such effects. Our Random Forest results
confirm that tree-level independence already yields high stability
(Section~\ref{sec:discussion}), validating the principle on a model
family that is independent by construction. Future work will explore
whether the mechanism extends to neural networks (where gradient-based
optimization creates different but potentially analogous path
dependencies) and whether partial orders on feature importance can
replace point rankings as a more robust representation of explanation
structure.

The empirical equivalence between \dash{} and Stochastic Retrain---and
the monotonic degradation of dependent methods with
$\rhovar$---is consistent with a formal impossibility result we
establish in companion work \citep{caraker2026impossibility}: no
single-model feature ranking can simultaneously be faithful (reflect
the model's attributions), stable (consistent across equivalent models),
and complete (rank all feature pairs) when features are collinear.
This trilemma holds for any attribution method applied to any model
class exhibiting the Rashomon property under correlation---not only
gradient boosting. The present work provides the empirical foundation
and constructive resolution; the companion theorem, mechanically
verified in the Lean~4 proof assistant, proves that ensemble-based
approaches like \dash{} are not merely effective but mathematically
necessary for achieving both stability and equity.

The neural network case warrants particular attention. Attribution
instability under feature collinearity is well-documented for NNs:
small input perturbations produce large ranking changes
\citep{ghorbani2019interpretation}, and independently trained models
achieving indistinguishable test loss can produce divergent Shapley
rankings \citep{damour2020underspecification}---a direct analog of
GBDT first-mover bias operating through initialization-driven symmetry
breaking rather than residual sequencing. The Rashomon set framework
\citep{fisher2019all, semenova2022existence} predicts that averaging
over independent initializations should cancel this bias by the same
argument that motivates \dash{}: the ensemble mean converges toward
the attribution implied by the basin center of mass, not any single
sample's idiosyncratic credit assignment. However, two differences
temper this prediction. First, NNs possess a wider Rashomon set under
high collinearity \citep{damour2020underspecification}, likely
requiring larger $K$ for equivalent stability gains. Second,
gradient-based attribution methods (Integrated Gradients, GradSHAP)
introduce baseline-sensitivity variance absent in TreeSHAP
\citep{alvarez2018towards}, confounding attribution instability with
method instability in any NN experiment. A careful extension of \dash{}
to neural architectures would need to control for both factors---a
promising but non-trivial direction. A concrete near-term extension
is stable feature interaction estimation: averaging TreeSHAP interaction
tensors across the \dash{} ensemble would provide stable pairwise
interaction rankings under multicollinearity, a capability that no
single-model workflow can offer.

\section{Conclusion}
\label{sec:conclusion}

We have investigated first-mover bias---the path-dependent concentration of
feature importance associated with sequential residual fitting in gradient
boosting---as a specific mechanistic contributor to SHAP instability under
multicollinearity. Three lines of evidence support this finding:

\begin{enumerate}
  \item The Large Single Model, which maximizes sequential dependency,
    produces the poorest attribution reproducibility of any method tested---worse
    than the standard single-best workflow---despite matching \dash{}'s
    total tree count and feature restriction. This provides strong evidence that
    sequential dependency, not model capacity, is a major driver of instability.
  \item \dash{} and Stochastic Retrain achieve equivalent
    stability ($0.977$ at $\rhovar = 0.9$; accuracy $d = +0.05$,
    equity $d = -0.13$, both n.s.; TOST confirmed),
    despite differing in every design choice except model independence.
    This supports the view that independence between explained models is the
    operative mechanism. \dash{}'s diversity-aware selection achieves this
    with $K_{\text{eff}} \approx 12$ models (vs.\ $K = 30$ for SR) and
    provides significantly more equitable attribution spreading within
    correlated groups ($p < 0.001$ vs.\ Random Selection at every
    $\rhovar$ level).
  \item The effect scales with correlation: dependent methods degrade
    monotonically as $\rhovar$ increases, while independent methods
    remain flat. This matches the mechanistic prediction.
\end{enumerate}

\dash{} (Diversified Aggregation of SHAP) operationalizes the
independence principle at two layers: the \emph{core operation}---averaging
attributions across independent models, proved to be the minimum-variance
unbiased estimator \citep{caraker2026impossibility}---and the
\emph{pipeline}, which adds forced feature restriction, diversity-aware
model selection, and two diagnostic tools (the Feature Stability
Index and Importance-Stability Plot) that detect first-mover bias
without ground truth. On real-world datasets, \dash{} improves stability
from $0.339$ to $0.925$ (Breast Cancer), from $0.840$ to $0.964$
(Superconductor), and from $0.97$ to $0.98$ (California Housing).

The code, data, and reproducible benchmarks are publicly available at
\url{https://github.com/DrakeCaraker/dash-shap}.

\section*{Reproducibility Statement}

All code, data generators, hyperparameter configurations, and evaluation
metrics are publicly available at
\url{https://github.com/DrakeCaraker/dash-shap}. The complete benchmark
can be reproduced via \texttt{python run\_experiments\_parallel.py} with fixed
random seeds ($\text{SEED} = 42$). The authoritative interactive
notebook \texttt{notebooks/demo\_benchmark\_7\_parallel.ipynb} provides a
checkpointed walkthrough of all experiments. The canonical results were
produced on AWS SageMaker (\texttt{ml.g5.16xlarge}, 64~vCPU, 248~GB~RAM);
hardware-dependent timing results (Table~\ref{tab:cost}) were measured on
Apple M-series silicon, single node, and are reported for relative comparison only.

\section*{Acknowledgments}

The authors thank the open-source XGBoost and SHAP communities for
providing the computational infrastructure on which this work is built.

\paragraph{License.}
This work is licensed under the Creative Commons Attribution 4.0
International License (CC-BY 4.0). To view a copy of this license, visit
\url{https://creativecommons.org/licenses/by/4.0/}.

\bibliographystyle{plainnat}

\appendix

\section{Algorithm Pseudocode}
\label{app:algorithm}

\begin{algorithm}[htbp]
\caption{DASH Pipeline}
\label{alg:dash}
\begin{algorithmic}[1]
\REQUIRE Training data $(X_{\text{train}}, y_{\text{train}})$, validation
  data $(X_{\text{val}}, y_{\text{val}})$, reference data $X_{\text{ref}}$,
  population size $M$, max ensemble size $K$, performance threshold
  $\varepsilon$, diversity threshold $\delta$, search space $\Theta$
\ENSURE Consensus SHAP matrix $\bar{\Phi}$, diagnostics (FSI, IS Plot)

\STATE \textbf{Stage 1: Population Generation}
\FOR{$i = 1$ to $M$}
  \STATE Sample $\theta_i \sim \text{Uniform}(\Theta)$
  \STATE Train $f_i \gets \text{XGBoost}(X_{\text{train}}, y_{\text{train}};
    \theta_i, \text{seed}=i)$
  \STATE Evaluate $s_i \gets \text{score}(f_i, X_{\text{val}}, y_{\text{val}})$
\ENDFOR

\STATE \textbf{Stage 2: Performance Filtering}
\STATE $s^* \gets \max_i s_i$
\STATE $\mathcal{F} \gets \{i : |s_i - s^*| \leq \varepsilon\}$

\STATE \textbf{Stage 3: Diversity Selection}
\FOR{$i \in \mathcal{F}$}
  \STATE $\mathbf{v}_i \gets \text{gain\_importance}(f_i, X_{\text{ref}})$
\ENDFOR
\STATE $\mathcal{S} \gets \text{MaxMinSelect}(\{\mathbf{v}_i\}_{i \in \mathcal{F}},
  K, \delta)$

\STATE \textbf{Stage 4: Consensus SHAP}
\FOR{$i \in \mathcal{S}$}
  \STATE $\Phi^{(i)} \gets \text{TreeSHAP}(f_i, X_{\text{ref}})$
\ENDFOR
\STATE $\bar{\Phi} \gets \frac{1}{|\mathcal{S}|} \sum_{i \in \mathcal{S}}
  \Phi^{(i)}$

\STATE \textbf{Stage 5: Diagnostics}
\STATE $\bar{I}_j \gets \frac{1}{N'}\sum_n |\bar{\Phi}_{nj}|$ for each
  feature $j$
\STATE $\text{FSI}_j \gets \bar{\sigma}_j / (\bar{I}_j + \epsilon_0)$ for
  each feature $j$
\RETURN $\bar{\Phi}$, FSI, IS Plot
\end{algorithmic}
\end{algorithm}

\section{Extended Results and Reproducibility}
\label{app:full_sweep}

Full per-repetition importance vectors, significance test results, and
ablation data are available in JSON format from the experimental runner.
Running \texttt{python run\_experiments\_parallel.py} reproduces all tables and
figures. The authoritative notebook
\texttt{notebooks/demo\_benchmark\_7\_parallel.ipynb} provides a checkpointed
walkthrough of all experiments with intermediate results cached for
rapid re-execution.

All code, data generators, and benchmark infrastructure are publicly
available at \url{https://github.com/DrakeCaraker/dash-shap}.

\section{Ablation Studies and Computational Cost}
\label{app:ablation}

\paragraph{Epsilon sensitivity.}
\dash{} is remarkably robust to the performance filter threshold
$\varepsilon$ (Table~\ref{tab:epsilon}). Across a $3\times$ range of
$\varepsilon$ values ($0.03$ to $0.10$), stability varies by $< 0.005$.
The effective ensemble size $K_{\text{eff}}$ scales with $\varepsilon$
(4.0 to 16.2), but performance plateaus early.

\begin{table}[htbp]
\centering
\caption{Epsilon sensitivity at $\rhovar = 0.9$ (50 repetitions).
Performance is robust across a $3\times$ range.}
\label{tab:epsilon}
\begin{tabular}{cccccc}
\toprule
$\varepsilon$ & Models Passing & $K_{\text{eff}}$ & Stability & Accuracy & Equity \\
\midrule
0.03 & 9.3  & 4.0 $\pm$ 1.4  & 0.9759 & 0.9876 & 0.183 \\
0.05 & 21.0  & 6.6 $\pm$ 1.9 & 0.9764 & 0.9879 & 0.178 \\
0.08 & 46.1 & 11.9 $\pm$ 3.0 & 0.9767 & 0.9881 & 0.175 \\
0.10 & 63.0 & 16.1 $\pm$ 3.6 & 0.9774 & 0.9884 & 0.173 \\
\bottomrule
\end{tabular}
\end{table}

\paragraph{Population size ablation.}
Stability is robust across population sizes $M$:
$M = 50$ ($0.9728$) $\to$ $M = 100$ ($0.9759$)
$\to$ $M = 200$ ($0.9767$) $\to$ $M = 500$ ($0.9777$). Performance
is effectively invariant to population size (within 0.001 across
$M \in \{50, 100, 200, 500\}$). We use $M = 200$ as a conservative default.

\begin{figure}[H]
\centering
\includegraphics[width=\textwidth,alt={Four-panel ablation sensitivity plot showing stability vs M, K, epsilon, and delta at three correlation levels}]{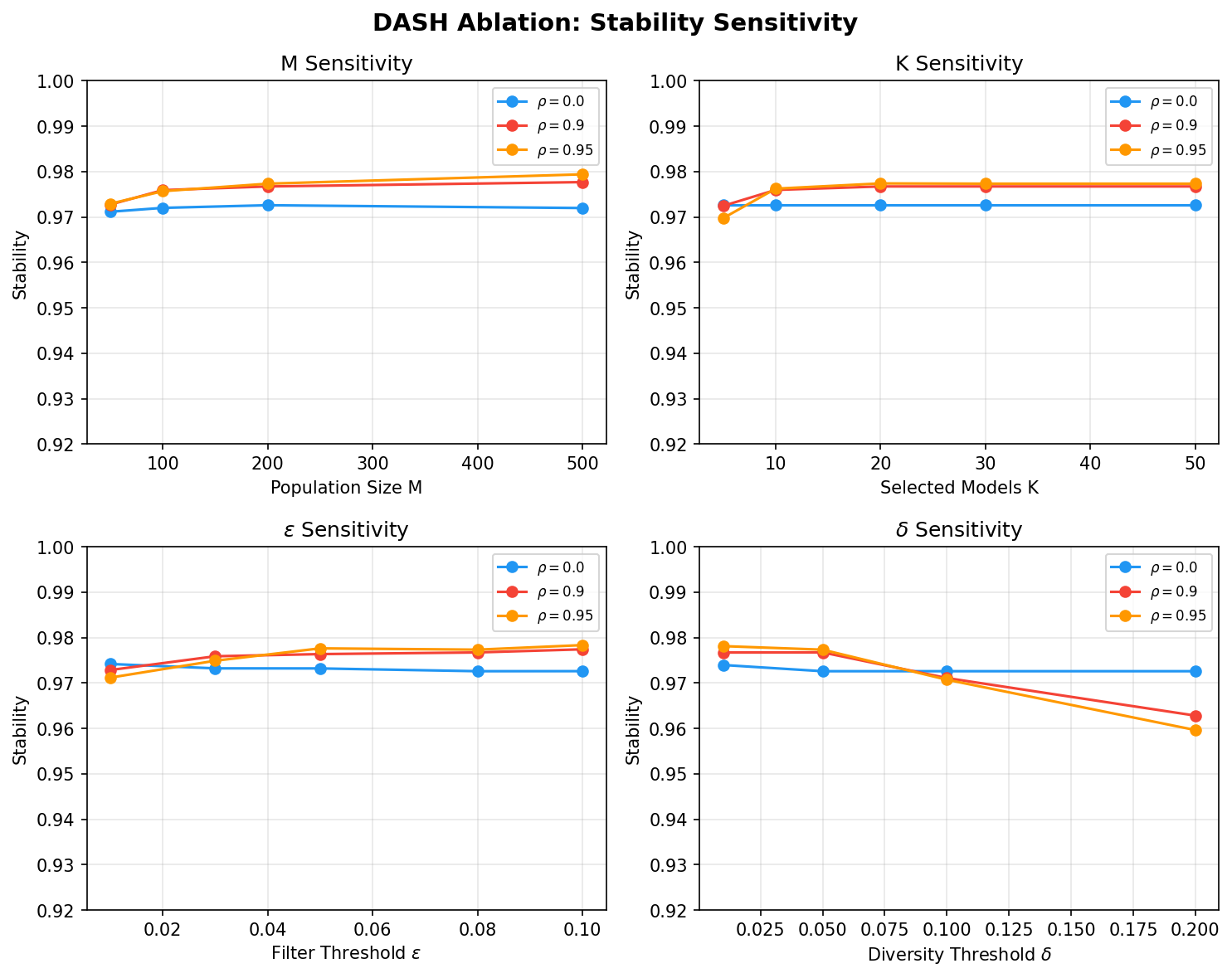}
\caption{Ablation sensitivity: stability as a function of each \dash{}
hyperparameter at three correlation levels
($\rhovar \in \{0.0, 0.9, 0.95\}$).
\textbf{Top left}: Population size $M$---stability is effectively
invariant ($\Delta < 0.005$ across $M \in \{50, 100, 200, 500\}$).
\textbf{Top right}: Selected models $K$---saturates at $K \approx 20$.
\textbf{Bottom left}: Filter threshold $\varepsilon$---robust across
a $3\times$ range ($0.03$--$0.10$).
\textbf{Bottom right}: Diversity threshold $\delta$---sensitive above
$0.05$; the default $\delta = 0.05$ is at the elbow.}
\label{fig:ablation}
\end{figure}

\paragraph{Computational cost.}
Table~\ref{tab:cost} reports wall-clock time for each method at
$\rhovar = 0.9$ (50~repetitions, single-threaded timing).
\dash{}'s cost is dominated by training $M = 200$ models and computing
$K \leq 30$ TreeSHAP explanations.

\begin{table}[H]
\centering
\caption{Computational cost at $\rhovar = 0.9$ (50 repetitions).
Wall-clock times are hardware-dependent (Apple M-series, single thread)
and reported for relative comparison; absolute times on the SageMaker
\texttt{ml.g5.16xlarge} instance are $\sim$12$\times$ higher due to
\texttt{nthread=1} per-rep parallelism.}
\label{tab:cost}
\small
\begin{tabular}{lcccc}
\toprule
\textbf{Method} & \textbf{Models} & \textbf{SHAP Evals} & \textbf{Per-rep (s)} & \textbf{Ratio} \\
\midrule
Large Single Model & 1  & 1  & \phantom{00}6.4 & $0.1\times$ \\
Single Best       & 30  & 1  & \phantom{0}43.5 & $1.0\times$ \\
LSM (Tuned)       & 1   & 1  & \phantom{0}88.9 & $2.0\times$ \\
\dash{} (MaxMin)  & 200 & $K_{\text{eff}}$ & 140.3 & $3.2\times$ \\
Stochastic Retrain & 30 & 30 & 233.5 & $5.4\times$ \\
Single Best ($M{=}200$)  & 200 & 1  & 248.8 & $5.7\times$ \\
Random Selection  & 200 & $K_{\text{eff}}$ & 287.2 & $6.6\times$ \\
\bottomrule
\end{tabular}
\smallskip

\noindent\textit{Note:} \dash{}'s diversity selection typically terminates
before reaching $K_{\max} = 30$ (the minimum-distance threshold $\delta$
stops selection early), yielding $K_{\text{eff}} \approx 10$--$15$ SHAP
evaluations at $\varepsilon = 0.08$. Random Selection always selects
$K = 30$ models, requiring roughly twice as many SHAP computations per
repetition. Stochastic Retrain similarly computes SHAP for all $K = 30$
models, explaining its higher cost relative to \dash{}.
\end{table}

\section{Pre-Specified Success Criteria}
\label{app:criteria}

We pre-specified eleven pass/fail criteria before running the final
benchmark (written into the experimental notebook prior to execution,
though not lodged with a formal pre-registration registry). These criteria test the paper's stated hypotheses under
favorable conditions (known-DGP synthetic data and datasets with
documented collinearity); adversarial or out-of-distribution stress
tests are deferred to the journal version. \dash{} passes all eleven:

\begin{samepage}
\begin{enumerate}
  \item \textbf{Stability wins (linear)}: DASH $>$ Single Best on $\geq$4/5
    $\rhovar$ levels $\to$ \textbf{PASS} (4/5)
  \item \textbf{Accuracy at $\rhovar = 0.9$}: DASH $\geq$ SB $\to$
    \textbf{PASS} (DASH $= 0.9879$ vs.\ SB $= 0.9784$)
  \item \textbf{Equity wins (linear)}: DASH $<$ Single Best CV on $\geq$4/5
    $\rhovar$ levels $\to$ \textbf{PASS} (5/5)
  \item \textbf{Safety control at $\rhovar = 0$}: No degradation vs.\
    baselines $\to$ \textbf{PASS} (gap $= 0.0003$)
  \item \textbf{$K_{\text{eff}}$ increases with $\varepsilon$}: monotonic
    $\to$ \textbf{PASS} ($4.0 \to 6.6 \to 11.9 \to 16.1$)
  \item \textbf{Nonlinear DGP}: DASH $>$ SB stability at $\rhovar = 0.9$
    $\to$ \textbf{PASS} (DASH $= 0.887$ vs.\ SB $= 0.811$)
  \item \textbf{Statistical significance}: $\geq$50\% of tests significant
    $\to$ \textbf{PASS} (17/26 Bonferroni, 15/26 Holm--Bonferroni)
  \item \textbf{Superconductor}: DASH stability $>$ SB $\to$ \textbf{PASS}
    ($0.964$ vs.\ $0.840$)
  \item \textbf{California Housing}: DASH stability $>$ SB $\to$ \textbf{PASS}
    ($0.978$ vs.\ $0.969$; $p = 0.063$ n.s.\ but positive direction)
  \item \textbf{Breast Cancer}: DASH stability $>$ SB $\to$ \textbf{PASS}
    ($0.925$ vs.\ $0.376$)
  \item \textbf{Variance decomposition}: DASH model-selection variance $<$ SB
    $\to$ \textbf{PASS} ($0.006$ vs.\ $0.023$)
\end{enumerate}
\end{samepage}

\section{Asymmetric Causal DGP Check}
\label{app:asymmetric}

\paragraph{Motivation.}
The linear DGP used in the main experiments assigns equal true importance
to all features within a correlated group, which means that \dash{}'s
equity metric (within-group CV) is evaluated in a regime where equity
and accuracy are perfectly aligned by construction.
A natural concern is whether \dash{} \emph{over-equalizes} when one
feature is causally active and its correlate is a passive proxy---that
is, does aggregation over independent models wash out a legitimate
importance asymmetry?

\paragraph{DGP.}
We use a two-feature asymmetric causal DGP:
\[
  y = 2 f_0 + \varepsilon, \quad
  f_1 = \rho f_0 + \sqrt{1 - \rho^2}\, z, \quad
  \varepsilon, z \sim \mathcal{N}(0, \sigma^2),
\]
where $f_0$ is causally active and $f_1$ is a passive correlate.
Ground-truth importance: $f_0 = 1.0$, $f_1 = 0.0$ (normalized).
We sweep $\rho \in \{0.5, 0.7, 0.9, 0.95\}$ with $N = 5{,}000$ and
$N_{\text{reps}} = 50$ repetitions.

\paragraph{Metrics.}
For each method we report:
\begin{itemize}
  \item \textbf{Stability}: mean pairwise Spearman correlation across
    repetitions (same as the main experiments).
  \item \textbf{Attribution bias}: $|\overline{\hat{\phi}}_0 - \text{true}(f_0)|$,
    the absolute deviation of the mean SHAP attribution for $f_0$ from its
    ground-truth importance (after normalizing attributions to sum to 1).
  \item \textbf{Passive leak}: mean attribution assigned to $f_1$
    (ground truth: 0.0); a proxy for over-equalization.
\end{itemize}

\paragraph{Results.}
Table~\ref{tab:asymmetric} reports results at $\rho = 0.9$ (the
primary evaluation condition in the main experiments).
\dash{} correctly preserves the causal asymmetry: all methods attribute
substantially more importance to the causal feature $f_0$ than the
passive correlate $f_1$. However, \dash{}'s passive leak (0.089) is
higher than Single Best's (0.068), reflecting a trade-off inherent to
ensemble averaging: some models in the ensemble attribute to $f_1$, and
their contributions are not zeroed out by averaging. Stochastic Retrain
shows a similar pattern (0.074). This trade-off increases with $\rho$
(DASH passive leak rises from 0.046 at $\rho{=}0.5$ to 0.173 at
$\rho{=}0.95$), and is the expected cost of the variance reduction
that \dash{} provides.

\begin{table}[htbp]
\centering
\caption{Asymmetric causal DGP at $\rho = 0.9$, 50 repetitions.
``Passive leak'' is mean attribution to the causally inert feature $f_1$
(ground truth: 0.0). Lower is better for both bias and passive leak.
All methods achieve stability = 1.000 (deterministic DGP).}
\label{tab:asymmetric}
\small
\begin{tabular}{lccc}
\toprule
\textbf{Method} & \textbf{Stability} & \textbf{Bias ($f_0$)} & \textbf{Passive leak ($f_1$)} \\
\midrule
Single Best & 1.000 & 0.068 & 0.068 \\
Stochastic Retrain & 1.000 & 0.074 & 0.074 \\
\textbf{DASH (MaxMin)} & 1.000 & 0.089 & 0.089 \\
Large Single Model & 1.000 & 0.084 & 0.084 \\
\bottomrule
\end{tabular}
\end{table}

\section{Two-Tree Analytical Example}
\label{app:twotree}

To build intuition for first-mover bias, we analyze the simplest possible
case: two-step linear boosting on correlated features. The companion
paper \citep{caraker2026impossibility} proves the formal result---the
attribution ratio is exactly $1/(1-\rhovar^2)$ for gradient boosting
under the proportionality axiom---and verifies it in Lean~4. Here we
provide an accessible derivation that isolates the core mechanism.

\paragraph{Setup.}
Let $A, B$ be two features with $\text{Cov}(A, B) = \rho$,
$\text{Var}(A) = \text{Var}(B) = 1$, and target $y = A + \varepsilon$.
Consider linear boosting with learning rate $\eta$: at each step we
fit $y$ (or its residual) by regressing on a single feature and updating
the model with the fitted value scaled by $\eta$.

\paragraph{Step~1.}
Suppose the first step selects feature $A$.  The fitted coefficient is
$\hat{\beta}_1 = \text{Cov}(y, A)/\text{Var}(A) = 1$, so the model
after step~1 is $f_1(x) = \eta A$.  The residual is
$r_1 = y - \eta A = (1 - \eta) A + \varepsilon$.

\paragraph{Step~2.}
Now consider the gain from each feature on the residual $r_1$:
\begin{align}
  \text{Gain}(A) &= \bigl[\text{Cov}(r_1, A)\bigr]^2 / \text{Var}(A) = (1 - \eta)^2, \\
  \text{Gain}(B) &= \bigl[\text{Cov}(r_1, B)\bigr]^2 / \text{Var}(B) = \rho^2(1 - \eta)^2.
\end{align}
Since $\rho < 1$, feature $A$ has strictly higher gain on the residual:
$\text{Gain}(A) / \text{Gain}(B) = 1/\rho^2$.  The second step selects
$A$ again.  After two steps, the model is $f_2(x) = \eta(2 - \eta) A$:
\emph{only feature $A$ is used}, and all SHAP credit goes to $A$.

\paragraph{From gain bias to concentration.}
The gain ratio $1/\rho^2 \approx 1.23$ at $\rhovar = 0.9$ demonstrates
the mechanism: after one step, the residual structure encodes which
feature was used, creating a bias toward re-selecting that feature.

This simple linear model is deliberately minimal.  Two features and
linear gain are sufficient to isolate the residual-bias mechanism, but
they do not capture the full concentration dynamics of tree-based
boosting.  In particular, when the DGP is symmetric
($y = \beta \bar{z}_g + \varepsilon$ with $\bar{z}_g = (A + B)/2$,
the paper's main setting), linear boosting alternates between $A$
and~$B$ at each step---the unused feature always has marginally higher
gain---so no net concentration arises in this idealization.

In actual XGBoost, three additional factors break this alternation and
enable the concentration observed in Figure~\ref{fig:t_scaling}:
(1)~threshold-based splits create nonlinear gain functions that do not
alternate as cleanly as linear regression gains;
(2)~multi-depth trees create interaction structure where a feature used
at the root influences which features are useful at child nodes; and
(3)~\texttt{colsample\_bytree} restricts the available feature set per
tree, so the feature that accumulates slightly more early splits
(by chance) builds a residual advantage that compounds over hundreds
of trees.  The interplay of these stochastic and structural effects is
what produces the monotonically growing concentration in
Figure~\ref{fig:t_scaling}---the linear gain-bias derived above provides
the seed, and tree-specific dynamics amplify it.

\paragraph{Independence resolves the compounding.}
Because independently trained models make their first-mover selections
independently, the compounding runs in different directions for different
models.  When their SHAP vectors are averaged:
\begin{equation}
  \bar{\phi}_A \approx \tfrac{1}{K}\sum_{k=1}^{K} \phi_A^{(k)},
\end{equation}
models that concentrated on $A$ and models that concentrated on $B$
cancel each other's arbitrary choices. With $K$ independent models, the
variance of the consensus importance due to first-mover effects decays
as $O(1/K)$, explaining the stability plateau at $K \approx 20$
observed in the $K$-sweep experiment.

This example identifies the residual gain-bias seed ($1/\rho^2$)
and the cancellation principle ($O(1/K)$ variance decay).
Figure~\ref{fig:t_scaling} confirms the empirical predictions:
(1)~concentration grows monotonically with $T$ for a single sequential
model, and (2)~independent ensembles remain flat regardless of
per-model tree count.

\section{Attribution-Agnostic Interface Validation}
\label{app:lime_demo}

The \texttt{fit\_from\_attributions()} interface (contribution~5 in
Section~\ref{sec:introduction}) decouples \dash{}'s aggregation stages
(filtering, diversity selection, consensus, diagnostics) from the
attribution method. We validate this claim by applying \dash{} to LIME
\citep{ribeiro2016lime} attributions on the Breast Cancer dataset.

\paragraph{Procedure.}
We train $M = 30$ XGBoost classifiers with randomly sampled
hyperparameters (same search space as the main experiments,
\texttt{colsample\_bytree} $\in \{0.2, 0.3, 0.4, 0.5\}$).
For each model, we compute LIME attributions for $N' = 30$ observations
from the held-out explain set, producing an $(M, N', P)$ attribution
tensor. We then call
\texttt{pipe.fit\_from\_attributions(attributions, val\_scores)} with
$K = 10$ and relative $\varepsilon = 0.05$.

\paragraph{Results.}
\dash{} successfully executes all four post-population stages on the
LIME tensor. The consensus top features---\texttt{worst concave points}
(0.104), \texttt{worst texture} (0.102), \texttt{worst area} (0.086),
\texttt{worst perimeter} (0.073)---draw from the same clinically
relevant feature families (concavity, area, perimeter) as the
TreeSHAP-based rankings in Table~\ref{tab:realworld}, confirming that
the diagnostic framework generalizes across attribution methods.
MaxMin diversity selection yields $K_{\text{eff}} = 3$ (low because all
30 models achieve identical validation accuracy on this dataset,
limiting diversity in the filtered pool).

This demonstrates that \dash{}'s stages 2--5 can operate on any
$(M, N', P)$ attribution tensor, regardless of whether it was produced
by TreeSHAP, LIME, Integrated Gradients, or any other feature-level
attribution method. The independence principle---that averaging over
independently trained models cancels arbitrary attribution
choices---applies to any attribution method that is sensitive to model
specification, not only TreeSHAP.

\end{document}